\documentclass[twoside,11pt]{article}
\usepackage[utf8]{inputenc}
\usepackage{jair, theapa, rawfonts}
\usepackage[nolist]{acronym}
\usepackage{graphicx}
\usepackage{framed}
\usepackage{dialogue}
\usepackage{booktabs}
\usepackage{xcolor}
\usepackage{arydshln}

\newcommand{\todo}[1]{} 
\renewcommand{\todo}[1]{{\noindent \color{red} \textbf{TODO: } {#1}}} 

\ShortHeadings{Deep Dialog Act Recognition}
{Ribeiro, Ribeiro, \& Martins de Matos}
\firstpageno{1}

\begin{document}

\title{Deep Dialog Act Recognition using Multiple Token, Segment, and Context Information Representations}

\author{\name Eug\'{e}nio Ribeiro \email eugenio.ribeiro@inesc-id.pt \\
    \addr L$^2$F - Spoken Language Systems Laboratory, INESC-ID Lisboa \\
    Instituto Superior T\'{e}cnico, Universidade de Lisboa, Portugal
    \AND
    \name Ricardo Ribeiro \email ricardo.ribeiro@inesc-id.pt \\
    \addr L$^2$F - Spoken Language Systems Laboratory, INESC-ID Lisboa \\
    Instituto Universit\'{a}rio de Lisboa (ISCTE-IUL), Portugal
    \AND
    \name David Martins de Matos \email david.matos@inesc-id.pt \\
    \addr L$^2$F - Spoken Language Systems Laboratory, INESC-ID Lisboa \\
    Instituto Superior T\'{e}cnico, Universidade de Lisboa, Portugal
}


\maketitle

%
%

\begin{acronym}
    \acro{AL}{Active Learning}
    \acro{ASR}{Automatic Speech Recognition}
    \acro{CBOW}{Continuous Bag of Words}
    \acro{CNN}{Convolutional Neural Network}
    \acro{DNN}{Deep Neural Network}
    \acro{DRLM}{Discourse Relation Language Model}
    \acro{DSTC4}{Dialog State Tracking Challenge 4}
    \acro{FCT}{Funda\c{c}\~{a}o para a Ci\^{e}ncia e a Tecnologia}
    \acro{GloVe}{Global Vectors for Word Representation}
    \acro{GPU}{Graphics Processing Unit}
    \acro{GRU}{Gated Recurrent Unit}
    \acro{HCNN}{Hierarchical Convolutional Neural Network}
    \acro{HMM}{Hidden Markov Model}
    \acro{k-NN}{k-Nearest Neighbors}
    \acro{LSTM}{Long Short-Term Memory}
    \acro{LVRNN}{Latent Variable Recurrent Neural Network}
    \acro{MRDA}{ICSI Meeting Recorder Dialog Act Corpus}
    \acro{NER}{Named Entity Recognition}
    \acro{NLP}{Natural Language Processing}
    \acro{NN}{Neural Network}
    \acro{OOV}{out of vocabulary}
    \acro{POS}{Part of Speech}
    \acro{QA}{Question Answering}
    \acro{ReLU}{Rectified Linear Unit}
    \acro{RCNN}{Recurrent Convolutional Neural Network}
    \acro{RNN}{Recurrent Neural Network}
    \acro{RNNLM}{Recurrent Neural Network Language Model}
    \acro{SVM}{Support Vector Machine}
    \acro{SwDA}{Switchboard Dialog Act Corpus}
    \acro{WER}{Word Error Rate}
\end{acronym}

%
%
%
%

\begin{abstract}

Automatic dialog act recognition is a task that has been widely explored over the years. In recent works, most approaches to the task explored different deep neural network architectures to combine the representations of the words in a segment and generate a segment representation that provides cues for intention. In this study, we explore means to generate more informative segment representations, not only by exploring different network architectures, but also by considering different token representations, not only at the word level, but also at the character and functional levels. At the word level, in addition to the commonly used uncontextualized embeddings, we explore the use of contextualized representations, which are able to provide information concerning word sense and segment structure. Character-level tokenization is important to capture intention-related morphological aspects that cannot be captured at the word level. Finally, the functional level provides an abstraction from words, which shifts the focus to the structure of the segment. Additionally, we explore approaches to enrich the segment representation with context information from the history of the dialog, both in terms of the classifications of the surrounding segments and the turn-taking history. This kind of information has already been proved important for the disambiguation of dialog acts in previous studies. Nevertheless, we are able to capture additional information by considering a summary of the dialog history and a wider turn-taking context. By combining the best approaches at each step, we achieve performance results that surpass the previous state-of-the-art on generic dialog act recognition on both the \acf{SwDA} and the \acf{MRDA}, which are two of the most widely explored corpora for the task. Furthermore, by considering both past and future context, similarly to what happens in an annotation scenario, our approach achieves a performance similar to that of a human annotator on \ac{SwDA} and surpasses it on \ac{MRDA}. 

\end{abstract}

%
%

%
%

\section{Introduction}
\label{sec:introduction}

In order to interpret its conversational partners' utterances, it is valuable for a dialog system to identify the generic intention behind the uttered words, as it provides an important clue concerning how each segment should be interpreted. That intention is revealed by dialog acts, which are the minimal units of linguistic communication~\shortcite{Searle1969}. Similarly to many other \ac{NLP} tasks~\shortcite{Manning2015,Goldberg2016}, most of the recent approaches on dialog act recognition take advantage of different \ac{NN} architectures, especially based on \acp{RNN}~\shortcite{Schmidhuber1990} and \acp{CNN}~\shortcite{Fukushima1980}. Variations in the architecture aim to capture information concerning the multiple aspects that are relevant for the task. Considering the findings of previous studies on dialog act recognition, there are three main aspects to be explored. First, at what level should a segment be tokenized and how should those tokens be represented to provide relevant information concerning intention? Then, how can those token representations be combined to generate a representation of the whole segment, while keeping information about the original tokens and the relations between them? Finally, how can different kinds of context information, such as that provided by the surrounding segments or concerning the speaker, be represented and combined with the representation of the segment to achieve the best possible performance on the task?

Most \ac{NN}-based dialog act recognition approaches perform tokenization at the word-level and use uncontextualized pre-trained embeddings to represent the tokens. These word embeddings are typically trained on large corpora using embedding approaches which capture information concerning words that commonly appear together~\shortcite{Mikolov2013,Pennington2014}. However, they are unable to provide information concerning word function and segment structure, which is relevant for the identification of many dialog acts. Furthermore, in these models the representation of a word is the same for all its occurrences. Since the same word may have different meanings according to the context surrounding it, the performance on many \ac{NLP} tasks has been improved by using embedding approaches that generate contextualized word representations that are able to capture word-sense information~\shortcite{Peters2018,Devlin2018}. Additionally, in previous studies, we have shown that there are cues for intention at a sub-word level, both in lemmas and affixes, which cannot be captured using word-level tokenization~\shortcite{Ribeiro2018,Ribeiro2019} and require a complementary character-based approach to be captured. Finally, there are also cues for intention in word abstractions, such as syntactic units, especially when considering dialog acts that are related to the structure of the segment. However, in spite of having been used on other \ac{NLP} tasks, these abstractions have not been explored in previous studies on automatic dialog act recognition.

The generation of the segment representation is the aspect on which most variations between existing approaches occur. Multiple \ac{DNN} architectures have been explored to combine the token representations into a single segment representation that captures relevant information for the task. Of the two state-of-the-art approaches on dialog act recognition, one uses a deep stack of \acp{RNN} to capture long distance relations between tokens~\shortcite{Khanpour2016}, while the other uses multiple parallel temporal \acp{CNN} to capture relevant functional patterns with different length~\shortcite{Liu2017}. Although these approaches focus on capturing different information, both have been proved successful on the task. Thus, an approach able to capture both kinds of information is expected to outperform both of these approaches.

Finally, a dialog act is not only related to the words in a segment, but also to the whole dialog context. That is, the intention of a speaker is influenced by the dialog history, as well as the expectation of its future direction. An important aspect to consider is the turn-taking history, since a set of segments may have different intentions depending on who the speakers are. However, only speaker changes in relation to the previous segment have been explored in previous studies~\shortcite{Liu2017}. In terms of information from the surrounding segments, its influence, especially that of preceding segments, has been thoroughly explored in at least two studies~\shortcite{Ribeiro2015,Liu2017}. However, although both studies focused on studying the importance of the dependencies between segments, the context information from the preceding segments was provided in the form of a flattened sequence of their classifications, which does not capture those dependencies.

In this article, we study and compare different solutions concerning these three aspects of dialog act recognition. We focus on exploring different representation approaches at the three-levels {--} token, segment, and context {--} by considering and comparing those explored in previous studies on the task and then introducing other approaches which address their limitations. We report results of experiments on both the \acf{SwDA}~\shortcite{Jurafsky1997} and the \acf{MRDA}~\shortcite{Janin2003}, which are two of the largest English corpora containing dialogs annotated for dialog acts. These two corpora have been widely explored in previous studies on automatic dialog act recognition, especially \ac{SwDA}, which is the most explored corpus for the task. Thus, we are able to use the results achieved in those studies as performance baselines for comparison with those achieved in our experiments. 

The remainder of the article is organized as follows: in Section~\ref{sec:related}, we start by providing an overview of previous studies on dialog act recognition, with special focus on those that use \acp{DNN}. Then, in the three following sections, we discuss the approaches we use to represent information at each step, starting with token representation in Section~\ref{sec:token}, followed by segment representation in Section~\ref{sec:segment}, and context information representation in Section~\ref{sec:context}. Section~\ref{sec:setup} describes our experimental setup, including the used datasets, in Section~\ref{ssec:datasets}, the generic architecture used in our experiments, in Section~\ref{ssec:architecture}, implementation details, in Section~\ref{ssec:implementation}, and the evaluation approach, in Section~\ref{ssec:evaluation}. The results of our experiments are discussed and compared with those of previous studies in Section~\ref{sec:results}. Finally, Section~\ref{sec:conclusions} states the most important conclusions of this study and provides pointers for future work.

%
%

%
%

\section{Related Work}
\label{sec:related}

Automatic dialog act recognition is a task that has been widely explored over the years, using multiple classical machine learning approaches. The first approach on the most explored corpus for the task, \ac{SwDA}, relied on \acp{HMM}~\shortcite{Baum1966} using word n-grams as features~\shortcite{Stolcke2000}. This approach achieved 71.0\% accuracy when applied to the manual transcriptions and 64.8\% when applied to automatic transcriptions with 41\% \ac{WER} on the test set. Since then, many other classical machine learning approaches have been explored. For instance, \shortciteA{Rotaru2002} used the \ac{k-NN} algorithm~\shortcite{Cover1967}, with the distance between neighbors being measured as the number of common bigrams between segments. \shortciteA{Sridhar2009} used a maximum entropy model combining lexical, syntactic, and prosodic features with context information from the surrounding segments. \shortciteA{Webb2010} applied a classification approach guided by heuristics based on cue phrases, that is, phrases that are highly indicative of a particular dialog act. The article by \shortciteA{Kral2010} provides an overview of these approaches. The best results on the task without using \acp{DNN} were achieved using \acp{SVM}. First, \shortciteA{Gamback2011} used word n-grams, wh-words, punctuations, and context information from the preceding segments as features, together with an \ac{AL} approach to select the most informative subset of the training data. When considering the tag set with 42 labels, this approach achieved 76.5\% accuracy when performing 10-fold cross-validation on the \ac{SwDA} corpus. We were able to improve that result to 79.1\% in our study on the influence of context information on the task~\shortcite{Ribeiro2015}. To do so, instead of representing the preceding segments in the same manner as the current one, we simply provided their classification. This result was achieved when using the gold standard annotations. When using automatically predicted classifications, the performance dropped between 1.0 and 2.8 percentage points depending on the approach used to generate the classifications, but was still above that obtained using the approach by \shortciteA{Gamback2011}.

In recent years, similarly to other text classification tasks, such as news categorization and sentiment analysis~\shortcite{Kim2014,Conneau2017}, most studies on dialog act recognition take advantage of different \ac{NN} architectures. To our knowledge, the first of those studies was that by \shortciteA{Kalchbrenner2013}, which used a \ac{CNN}-based approach to generate segment representations from randomly initialized word embeddings. Then, it used an \ac{RNN}-based discourse model that combined the sequence of segment representations with speaker information and output the corresponding sequence of dialog acts. By limiting the discourse model to consider information from the two preceding segments only, this approach achieved 73.9\% accuracy on the \ac{SwDA} corpus. 

\shortciteA{Lee2016} compared the performance of a \ac{LSTM}~\shortcite{Hochreiter1997} unit against that of a \ac{CNN} to generate segment representations from pre-trained embeddings of its words. In order to generate the corresponding dialog act classifications, the segment representations were then fed to a 2-layer feed-forward network, in which the first layer normalized the representations and the second selected the class with highest probability. In their experiments, the \ac{CNN}-based approach consistently led to similar or better results than the \ac{LSTM}-based one. The architecture was also used to provide context information from up to two preceding segments at two levels. The first level refers to the concatenation of the representations of the preceding segments with that of the current segment before providing it to the feed-forward network. The second refers to the concatenation of the normalized representations before providing them to the output layer. This approach achieved 65.8\%, 84.6\%, and 71.4\% accuracy on the \ac{DSTC4}~\shortcite{Kim2016}, \ac{MRDA}~\shortcite{Janin2003}, and \ac{SwDA} corpora, respectively. However, the influence of context information varied across corpora.

\shortciteA{Ji2016} explored a combination of \ac{NN} architectures and probabilistic graphical models. They used a \ac{DRLM} that combined a \ac{RNNLM}~\shortcite{Mikolov2010} with a latent variable model over shallow discourse structure. The first models the sequence of words in the dialog, while the latter models the relations between adjacent segments which, in this context, are the dialog acts. Thus, the model can perform word prediction using discriminatively-trained vector representations while maintaining a probabilistic representation of a targeted linguistic element, such as the dialog act. In order to function as a dialog act classifier, the model was trained to maximize the conditional probability of a sequence of dialog acts given a sequence of segments, achieving 77.0\% accuracy on the \ac{SwDA} corpus.

\shortciteA{Tran2017a} used a hierarchical \ac{RNN} with an attentional mechanism to predict the sequence of dialog act classifications of a complete dialog. The model is hierarchical, since it includes an utterance-level \ac{RNN} to generate the representation of an utterance from its tokens and another to generate the sequence of dialog act labels from the sequence of utterance representations. The attentional mechanism is between the two, since it uses information from the dialog-level \ac{RNN} to identify the most important tokens in the current utterance and filter its representation. Using this approach, they achieved 74.5\% accuracy on \ac{SwDA} and 63.3\% on the MapTask corpus~\shortcite{Anderson1991}. Later, they were able to improve the performance on \ac{SwDA} to 75.6\% by propagating uncertainty information concerning the previous predictions~\shortcite{Tran2017c}. Additionally, they experimented with gated attention in the context of a generative model, achieving 74.2\% on \ac{SwDA} and 65.94\% on MapTask~\shortcite{Tran2017b}.

The previous studies explored the use of a single recurrent or convolutional layer to generate the segment representation from those of its words. However, the top performing approaches use multiple of those layers. On the one hand, \shortciteA{Khanpour2016} achieved their best results using a segment representation generated by concatenating the outputs of a stack of 10 \ac{LSTM} units at the last time step. This way, the model is able to capture long distance relations between tokens. On the convolutional side, \shortciteA{Liu2017} generated the segment representation by combining the outputs of three parallel \acp{CNN} with different context window sizes, in order to capture different functional patterns. In both cases, pre-trained word embeddings were used as input to the network. Overall, from the reported results, it is not possible to state which is the top performing segment representation approach since the evaluation was performed on different subsets of the \ac{SwDA} corpus. Still, \shortciteA{Khanpour2016} reported 73.9\% accuracy on the validation set and 80.1\% on the test set, while \shortciteA{Liu2017} reported 74.5\% and 76.9\% accuracy on the two sets used to evaluate their experiments. \shortciteA{Khanpour2016} also reported 86.8\% accuracy on the \ac{MRDA} corpus.

In addition to the \ac{CNN}-based segment representation approach, \shortciteA{Liu2017} also explored the use of context information. Information concerning turn-taking was provided as a flag stating whether the speaker changed in relation to the previous segment and concatenated to the segment representation. In order to extract information from the surrounding segments, they explored the use of discourse models, as well as of approaches that concatenated the context information directly to the segment representation. The discourse models transform the model into a hierarchical one by generating a sequence of dialog act classifications from the sequence of segment representations. Thus, when predicting the classification of a segment, the surrounding ones are also taken into account. However, when the discourse model is based on a \ac{CNN} or on a bidirectional \ac{LSTM} unit, it considers information from future segments, which is not available to a dialog system. Still, even when relying on future information, the approaches based on discourse models performed worse than those that concatenated the context information directly to the segment representation. Concerning the latter, providing that information in the form of the classification of the surrounding segments led to better results than using their words, even when those classifications were obtained automatically. This conclusion is in line with what we had shown in our previous study using \acp{SVM}~\shortcite{Ribeiro2015}. Furthermore, both studies have shown that, as expected, the first preceding segment is the most important and that the influence decays with the distance. Using the setup with gold standard labels from three preceding segments, the results achieved by \shortciteA{Liu2017} on the two sets used to evaluate the approach improved to 79.6\% and 81.8\%, respectively.

It is important to make some remarks concerning tokenization and token representation. In all the previously described studies, tokenization was performed at the word level. Furthermore, with the exception of the first study~\shortcite{Kalchbrenner2013}, which used randomly initialized embeddings, and possibly those by \shortciteA{Tran2017a}, for which the embedding approach was not disclosed, the representation of those words was given by pre-trained embeddings. \shortciteA{Khanpour2016} compared the performance when using Word2Vec~\shortcite{Mikolov2013} and \ac{GloVe}~\shortcite{Pennington2014} embeddings trained on multiple corpora. Although both embedding approaches capture information concerning words that commonly appear together, the best results were achieved using Word2Vec embeddings. In terms of dimensionality, that study compared embedding spaces with 75, 150, and 300 dimensions. The best results were achieved when using 150-dimensional embeddings. However, 200-dimensional embeddings were used in other studies~\shortcite{Lee2016,Liu2017}, which was not one of the compared values.

In two previous studies~\shortcite{Ribeiro2018,Ribeiro2019} we have shown that there are also important cues for intention at a sub-word level which can only be captured when using a finer-grained tokenization, such as at the character-level. Those cues mostly refer to aspects concerning the morphology of words, such as lemmas and affixes. To capture that information, we adapted the \ac{CNN}-based segment representation approach by \shortciteA{Liu2017} to use characters instead of words as tokens. This way, we were able to explore context windows of different sizes around each token to capture those different morphological aspects. Our best results were achieved when using three parallel \acp{CNN} with window sizes three, five, and seven, which are able to capture affixes, lemmas, and inter-word relations, respectively. We have performed experiments on corpora in multiple languages and shown that the character-level approach is always able to capture important information and that its importance increases with the level of inflection of the language. Concerning the English corpora, when using information from the current segment only, we achieved 76.8\% and 73.2\% accuracy on the validation and test sets of the \ac{SwDA} corpus, respectively, and 56.9\% when performing 5-fold cross-validation on the LEGO corpus~\shortcite{Schmitt2012}. These results are in line with those of the word-level approach. However, the combination of the two levels improved the results to 78.0\%, 74.0\%, and 57.9\%, respectively, which shows that character- and word-level tokenizations provide complementary information. Finally, by providing context information from the preceding segments to the combined approach, we achieved 82.0\% accuracy on the validation set and 79.0\% on the test set of the \ac{SwDA} corpus and 87.2\% on the LEGO corpus.

%
%

%
%

\section{Token Representation}
\label{sec:token}

A segment is formed by multiple constituents or tokens. Thus, its representation is obtained through a combination of the representations of those constituents. As revealed by most studies described in Section~\ref{sec:related}, a segment is typically seen as a sequence of words. However, it can also be considered at other levels. For instance, from a finer-grained point of view, a segment can be seen as a sequence of characters. On the other hand, it can also be seen as a sequence of syntactic units, or other abstractions from characters or words. In this section, we discuss the information that can be captured by performing tokenization at each of these levels and describe the approaches on token representation that we explore in this study. Since we are focusing on the multiple steps of dialog act recognition approaches using \acp{DNN}, we only explore embedding representations, that is, those that represent a token as a vector of coordinates in a certain embedding space~\shortcite{Lavelli2004}.

\subsection{Word-Level Tokenization}

Since segments are typically seen as sequences of words, there has been extensive research on approaches to generate word-embedding representations that capture relevant word semantics. Most of such approaches, including those used in previous studies on dialog act recognition, generate embeddings for each word independently of its context~\shortcite{Collobert2011,Mikolov2013,Pennington2014,Levy2014,Bojanowski2017}. Thus, all the occurrences of a word have the same embedding representation. However, recently, state-of-the-art approaches on multiple \ac{NLP} tasks rely on the use of contextualized word representations~\shortcite{Peters2018,Radford2018,Devlin2018}. Such representations include information from the context in which a word appears and, thus, vary for different occurrences of the same word. Consequently, the representations generated for each word cannot be shared among datasets and, thus, contrarily to the uncontextualized counterparts, sets of pre-trained contextualized word embeddings are not available. In this study we explore the use of contextualized word embeddings for dialog act recognition and compare their performance with those of uncontextualized embeddings, both pre-trained and randomly initialized.

\subsubsection{Randomly Initialized Embeddings}
\label{sssec:random}

A simplistic approach to word representation is to randomly initialize the embedding representation of each word and let it adapt during the training phase. In this case, there is only one parameter that must be defined, which is the dimensionality of the embedding space. This dimensionality is the factor that defines the trade-off between ambiguity and sparseness. On the one hand, higher dimensionality leads to increased sparseness and memory requirements. On the other hand, lower dimensionality leads to ambiguity in the representation. However, up to a certain level, ambiguity is not necessarily harmful.

The major drawback of this approach is that the representations of the words become overfit to their occurrences in the training dialogs. This is not problematic and may even be beneficial in scenarios involving dialogs in the same domain or following a similar structure. However, that is not typically the case and the overfit representations impair the generalization capabilities of the trained classifiers. Furthermore, this approach is more susceptible to problems related to \ac{OOV} words.

\subsubsection{Uncontextualized Embeddings}

As previously stated, there has been extensive research on means to generate uncontextualized word-embedding representations. The most common of such approaches are the \ac{CBOW} model~\shortcite{Mikolov2013}, commonly known as Word2Vec, and \ac{GloVe}~\shortcite{Pennington2014}. \shortciteA{Khanpour2016} compared the performance of both approaches in the context of dialog act recognition and concluded that Word2Vec embeddings lead to better results than \ac{GloVe} embeddings. More recently, the FastText~\shortcite{Bojanowski2017} approach was devised, improving on the previous two by considering character-level information when generating the embedding representation of the words. Given its improved performance and since it was not used in previous studies on dialog act recognition, we explore it in this study.

All of the previous approaches generate representations based on the co-occurrence of adjacent words. However, many dialog acts are related to the structure of the segment and not just sequences of specific words. Thus, we also explore the dependency-based embedding approach by \shortciteA{Levy2014}, which also takes that structure into account and not only word co-occurrences. It does so by introducing information concerning syntactic dependencies between words in the embedding generation process. First, it generalizes the Word2Vec approach by allowing it to use arbitrary contexts instead of just those based on adjacent words. Then, it uses the syntactic contexts derived from automatically produced dependency parse-trees. That is, the embedding generated for a given word is based on the syntactic relations it participates in. Thus, embeddings generated using this approach are appropriate for the dialog act recognition task. 

An advantage of uncontextualized embedding approaches is that since every occurrence of a word has the same embeddding, the embeddings for each word can be pre-trained on large amounts of data. Thus, multiple sets of pre-trained embeddings using all the previous approaches are publicly available for many languages. Using such sets typically leads to results that generalize better, since they are trained on large amounts of data and not only on a reduced set focused on a particular domain. However, this also means that their generation does not take their future use on a specific task into account. In their study, \shortciteA{Liu2017} used pre-trained embeddings but let them adapt to the task during the training phase. We do not apply that approach in this study, since it requires the whole embedding architecture and its features to be reproduced, which can be impractical or even infeasible for some of the embedding approaches.

\subsubsection{Contextualized Embeddings}

When using uncontextualized embeddings, the representation of a word is fixed, independently of its sense and the context surrounding it. However, the meaning of a word and, consequently, how it contributes for the meaning of a segment, is highly influenced by both its sense and context~\shortcite{Camacho2018}. Thus, recently, there has been a research trend focusing on approaches to generate contextualized word representations~\shortcite{Peters2018,Radford2018,Devlin2018}. Such approaches include information from the context in which a word appears in its embedding representation. Thus, the representation of a word varies among its occurrences. However, while some approaches generate those variations through the modification of the uncontextualized representation~\shortcite{Peters2018}, others generate them independently~\shortcite{Radford2018,Devlin2018}. The drawback of using this kind of representations is that, since the representation varies among occurrences of the same word, one cannot use sets of pre-trained embeddings for each word. Still, the models for generating the embeddings can be pre-trained on large amounts of data and, thus, the overhead of using contextualized embeddings instead of their uncontextualized counterparts for each segment is just the inference time for generating the embeddings of its words. 

The use of contextualized embeddings has led to state-of-the-art results on multiple \ac{NLP} tasks, such as \ac{NER}, \ac{QA}, and textual entailment~\shortcite{Devlin2018}. Although they have not been applied to dialog act recognition before, they have the potential to outperform uncontextualized embeddings on the task, since a word may have different functions depending on its sense. Furthermore, the context surrounding each word reveals information concerning the structure of a segment. Thus, in this study, we explore the use of two contextualized embedding approaches. 

The first, ELMo~\shortcite{Peters2018}, is based on a stack of two bi-directional \acp{LSTM} and was the first approach used to generate contextualized representations. Its output provides a context-free representation of the word and context information at two levels, given by the output of each \ac{LSTM}. Experiments by its authors have shown that one of these levels provides information concerning the sense of the word, while the other is more related to syntax. Since these two levels modify the context-free representation and the value range of the latter is typically wider than those of the context levels, we use the sum of the three levels as the representation of each word.

The second contextualized representation approach, BERT~\shortcite{Devlin2018}, is based on the Transformer architecture~\shortcite{Vaswani2017} and currently leads to state-of-the-art results on multiple benchmark \ac{NLP} tasks. It uses token, segment, and positional embeddings as input and a variable number of self-attention layers in both its encoder and decoder. When provided a sequence of tokens, it outputs contextualized representations of each word, as well as a combined representation for the whole sequence. The latter can be connected directly to a classification layer, allowing the weights of the model to be fine-tuned to a specific task. However, that differs from the modular nature of the other experiments in our study. Furthermore, the model requires more resources than we have available for training and we would need to modify it for our experiments using segment context information. Thus, we rely on the contextualized representation for each word output by the last self-attention layer of the decoder, since it contains information from all the layers that precede it.

\subsection{Character-Level Tokenization}
\label{ssec:charlevel}

Although the majority of models used in text processing are word-based, there are also character-based models achieving high performance on text classification tasks, such as news article categorization and review rating~\shortcite{Zhang2015}. The main advantage of these models is that they are able to capture information concerning the morphology of words, such as their lemmas and affixes. Furthermore, they typically generalize better since they do not have to deal with the \ac{OOV} problem. On the other hand, when using an embedding space with the same dimensionality, character models are slower, since they need to process a larger amount of tokens.

Considering the task at hand, dialog acts are related to an intention which is transmitted in the form of words. Thus, the selected words are expected to have a function related to that intention. That function is typically related to the morphology of the words. For instance, there are cases, such as adverbs of manner and negatives, in which the function, and hence the intention, of a word is related to its affixes. On the other hand, there are cases in which considering multiple forms of the same lexeme independently does not provide additional information concerning intention and, thus, the lemma suffices.

In our previous studies~\shortcite{Ribeiro2018,Ribeiro2019}, we have shown that a character-level model that identifies character patterns of variable size is able to capture aspects concerning that morphological information which cannot be captured using word-level approaches and are relevant for dialog act recognition. Thus, we also explore character-level approaches in this study.

Instead of using a fixed one-hot representation of each character, we apply an embedding approach that randomly initializes the representation of each character and adapts it during the training phase. This way, the embedding representation captures relations between characters that commonly appear together.

\subsection{Functional-Level Tokenization}

There are some dialog acts that are more related to the structure of the segment or the functions of its words than to the presence of specific words or sets of words. Thus, it makes sense to explore tokenization at a level that abstracts those words. Considering what was said in the previous section concerning the importance of affixes, it is interesting to assess how important they are for the task. Removing the lemmas and keeping just the affixes is not common in text classification tasks. Thus, we explore the negative approach, that is, we replace the words with their lemmas and assess the loss of performance in comparison to the case when the original words are used. 

A more drastic approach is to replace the words by the corresponding morphosyntactic units in the form of \ac{POS} tags. This leads to a representation that captures the syntactic function of each word, as well as the structure of the segment when sets of words are considered together. We explore the use of both the coarse-grained Google Universal POS tag set~\shortcite{Petrov2012} and the fine-grained Penn Treebank tag set~\shortcite{Marcus1993}. While the first only covers the word type, the latter also includes information regarding morphological features of the words.

To represent the \ac{POS} tags we use an embedding approach similar to the described for characters in the previous section. That is, we use the dimensionality required for a one-hot representation of each \ac{POS} tag, but randomly initialize the representation and let it adapt during the training phase. This way, the representation is able to capture information concerning syntactic dependencies.

\subsection{Multi-Level Combinations}

Each level of tokenization allows us to capture information concerning different aspects of intention. However, each level has its limitations. For instance, although the approach that uses functional-level tokenization in the form of \ac{POS} tags is able to capture information that is relevant to identify dialog acts that are related to the structure of the segment or the generic function of the words, it discards word-specific information, necessary for the identification of other dialog acts. Thus, it should be paired with a word-level tokenization approach to combine their advantages. Furthermore, in our previous studies~\shortcite{Ribeiro2018,Ribeiro2019}, we have shown that while a character-level model is able to capture information which cannot be captured using word-level approaches, the reverse is also true. Thus, the best performance is achieved when both approaches are used in combination. Considering these advantages and limitations of using tokenization at each level and their complementarity, we also explore their combined use.

%
%

%
%

\section{Segment Representation}
\label{sec:segment}

Looking into the dialog act recognition approaches described in Section~\ref{sec:related}, we can see that they differ mainly in terms of the approach used for segment representation, by applying different network architectures to combine the representations of the tokens to generate that of the segment. Of the two top-performing segment representation approaches in the context of dialog act recognition when using word-level tokenization, one is based on \acp{RNN}~\shortcite{Khanpour2016} and the other is based on \acp{CNN}~\shortcite{Liu2017}. Both have their own advantages: the first focuses on capturing information from relevant sequences of tokens and the latter focuses on the context surrounding each token and, thus, on capturing relevant token patterns independently of where they occur in the segment. Concerning the task at hand, this is relevant since, among other aspects, while some dialog acts are distinguishable due to the order of the words in the segment (e.g. subject-auxiliary inversion in questions), others are distinguishable due to the presence of certain words or sequences of words independently of where they occur in the segment (e.g. greetings). Although \shortciteA{Lee2016} have shown that the segment representation generated by a \ac{CNN} leads to better results that the one generated by an \ac{LSTM}, they only considered a single of those layers. On the other hand, the approaches by \shortciteA{Khanpour2016} and \shortciteA{Liu2017} use multiple stacked or parallel layers. Thus, since the two approaches were not compared directly and were evaluated on different sets, we include both of them in our experiments. However, since each approach is expected to capture a different kind of information, but both kinds are relevant for the task, a third approach that merges at least some of the capabilities of the other two is expected to perform better on the task. \shortciteA{Lai2015} designed an approach based on \acp{RCNN} which aims at capturing both kinds of information. This approach achieved state-of-art performance on text classification tasks, such as document topic classification and movie review rating, but was not applied to dialog act recognition before. Thus, we explore its use in our experiments on the task.

In our previous studies using character-level tokenization~\shortcite{Ribeiro2018,Ribeiro2019}, we adapted the \ac{CNN}-based approach to deal with characters instead of words, in order to identify relevant character patterns of different sizes corresponding to intention-related morphological aspects. Since we must target those patterns in order to take advantage of the character-level tokenization, the \ac{RNN}- and \ac{RCNN}-based segment representation approaches are not appropriate. While the first considers information from additional tokens incrementally as it processes the sequence, the latter includes information from the whole sequence in the representation of each token. Thus, they do not allow us to focus on character patterns of specific sizes. Consequently, we only explore them when considering word-level tokenization. The characteristics of each approach and our adaptations of their original versions are described below. 

\subsection{\ac{RNN}-Based Segment Representation}
\label{ssec:rnn}

\begin{figure}[ht]
	\centering
	\includegraphics[scale=0.6]{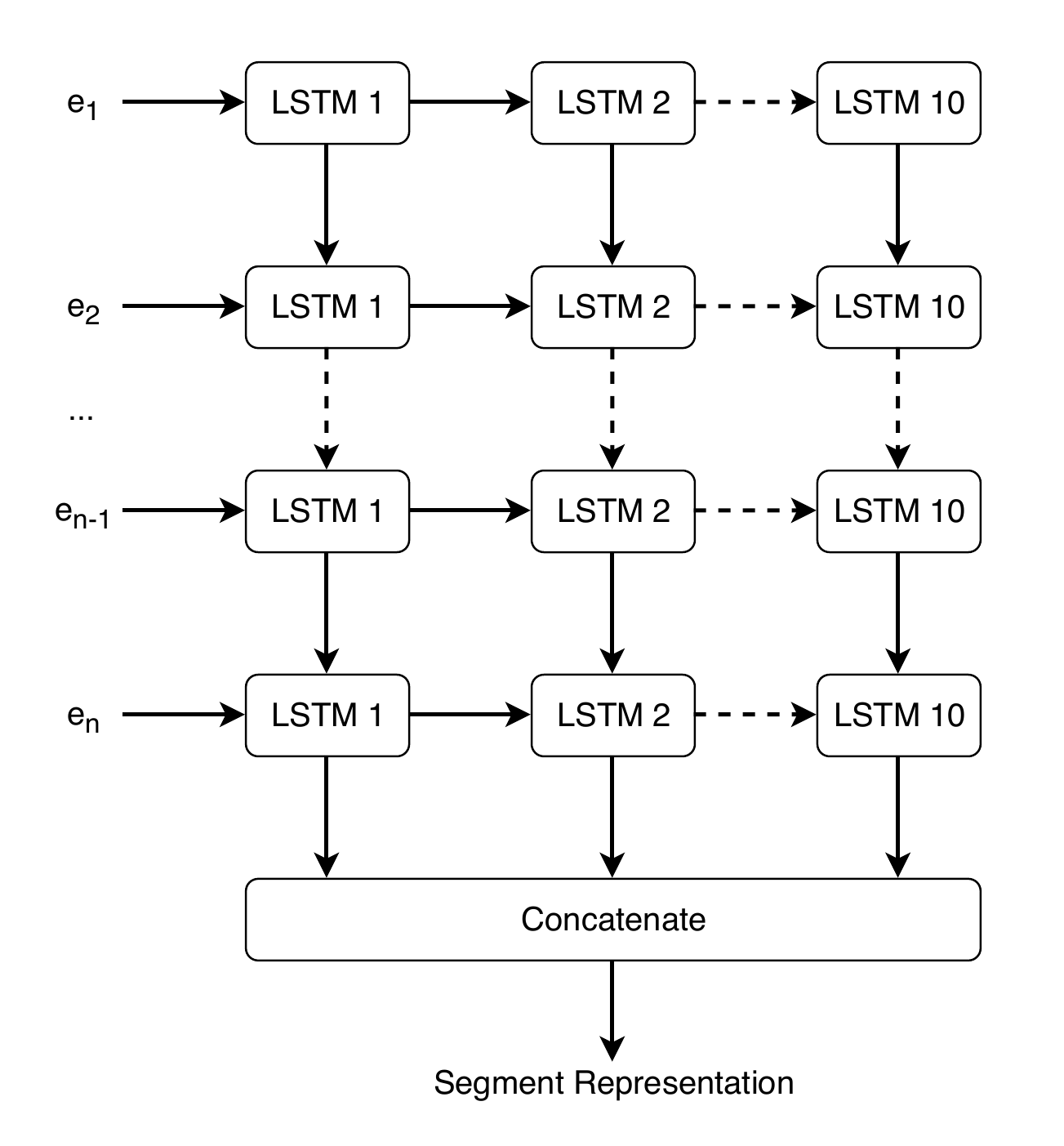}
	\caption{The \ac{RNN}-based segment representation approach. $e_i$ corresponds to the embedding representation of the $i$-th token.}
	\label{fig:rnn}
\end{figure}

The recurrent approach by \shortciteA{Khanpour2016} uses a stack of 10 \ac{LSTM} units. The segment representation is given by the concatenation of the outputs of all those units at the last time step, that is, after processing all the tokens in the segment. This process is shown in Figure~\ref{fig:rnn}. Using the output at the last time step instead of other pooling operation is appropriate in this scenario since the recurrent units process the tokens sequentially. Thus, the output at the last time step can be seen as a summary of the segment, containing information from all of its tokens. 

There are two adaptations that are expected to improve the performance of this approach. The first is replacing the \ac{LSTM} with \acp{GRU}~\shortcite{Chung2014}, in order to reduce resource consumption in terms of memory. The other is applying a bidirectional approach, that is, introducing another stack of \acp{LSTM} or \acp{GRU} which processes the sequence backwards. The representation of the segment is then given by the concatenation of the outputs of the two stacks. This way, it also includes information concerning the relations between each token and those that appear after it. However, resource consumption is highly increased.

\subsection{\ac{CNN}-Based Segment Representation}
\label{ssec:cnn}

\begin{figure}[ht]
	\centering
	\includegraphics[scale=0.6]{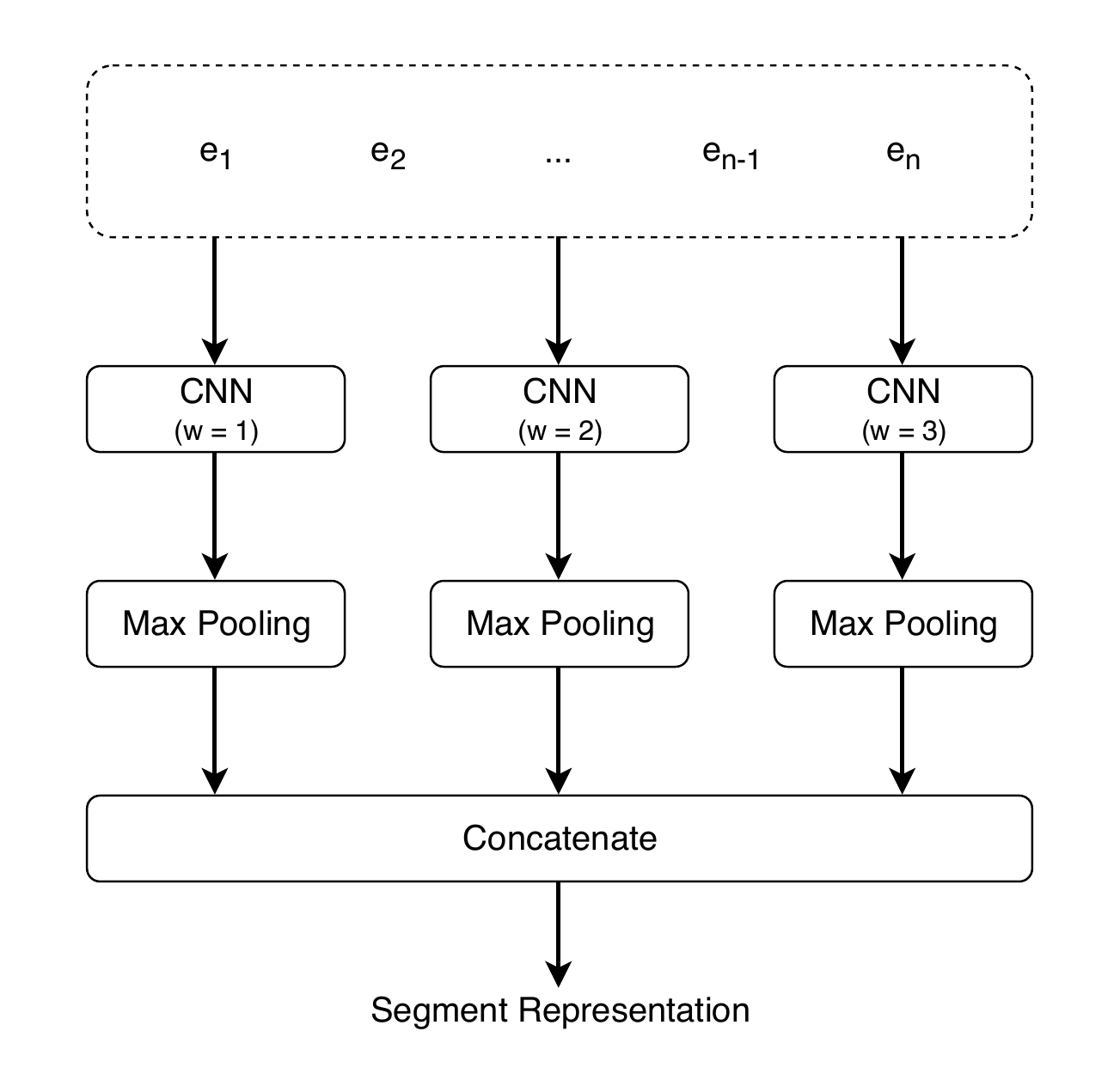}
	\caption{The \ac{CNN}-based segment representation approach. $e_i$ corresponds to the embedding representation of the $i$-th token. The parameter $w$ refers to the size of the context window.}
	\label{fig:cnn}
\end{figure}

The convolutional approach by \shortciteA{Liu2017} uses a set of parallel \acp{CNN} with different window size, each followed by a max-pooling operation. The segment representation is given by the concatenation of the results of the pooling operations. This way, the representation of the segment contains information concerning groups of tokens with different sizes. The process is shown in Figure~\ref{fig:cnn}.

To achieve the results presented in their paper, \shortciteA{Liu2017} used three \acp{CNN} with context window sizes between one and three. Thus, they consider each word individually, as well as word pairs and triples. In a previous study using the same architecture for different tasks, \shortciteA{Kim2014} used wider context windows, with sizes from three up to five. Considering the dialog act recognition task, the narrower windows are clearly relevant, since there are many dialog acts which can be identified by individual or small sets of words (e.g. greetings, confirmations, negations). However, the wider windows are able to capture relations between more distant tokens, which may also provide relevant information. Thus, it is interesting to assess whether that information can be used to improve the performance on the task. 

In our previous studies using character-level tokenization~\shortcite{Ribeiro2018,Ribeiro2019}, we used three \acp{CNN} with context window sizes three, five, and seven. These are able to capture small English affixes, longer affixes and lemmas, and long words and inter-word information, respectively. Thus, we use the same approach in this study when considering character-level tokenization.

\subsection{\ac{RCNN}-Based Segment Representation}

\begin{figure}[ht]
	\centering
	\includegraphics[scale=0.6]{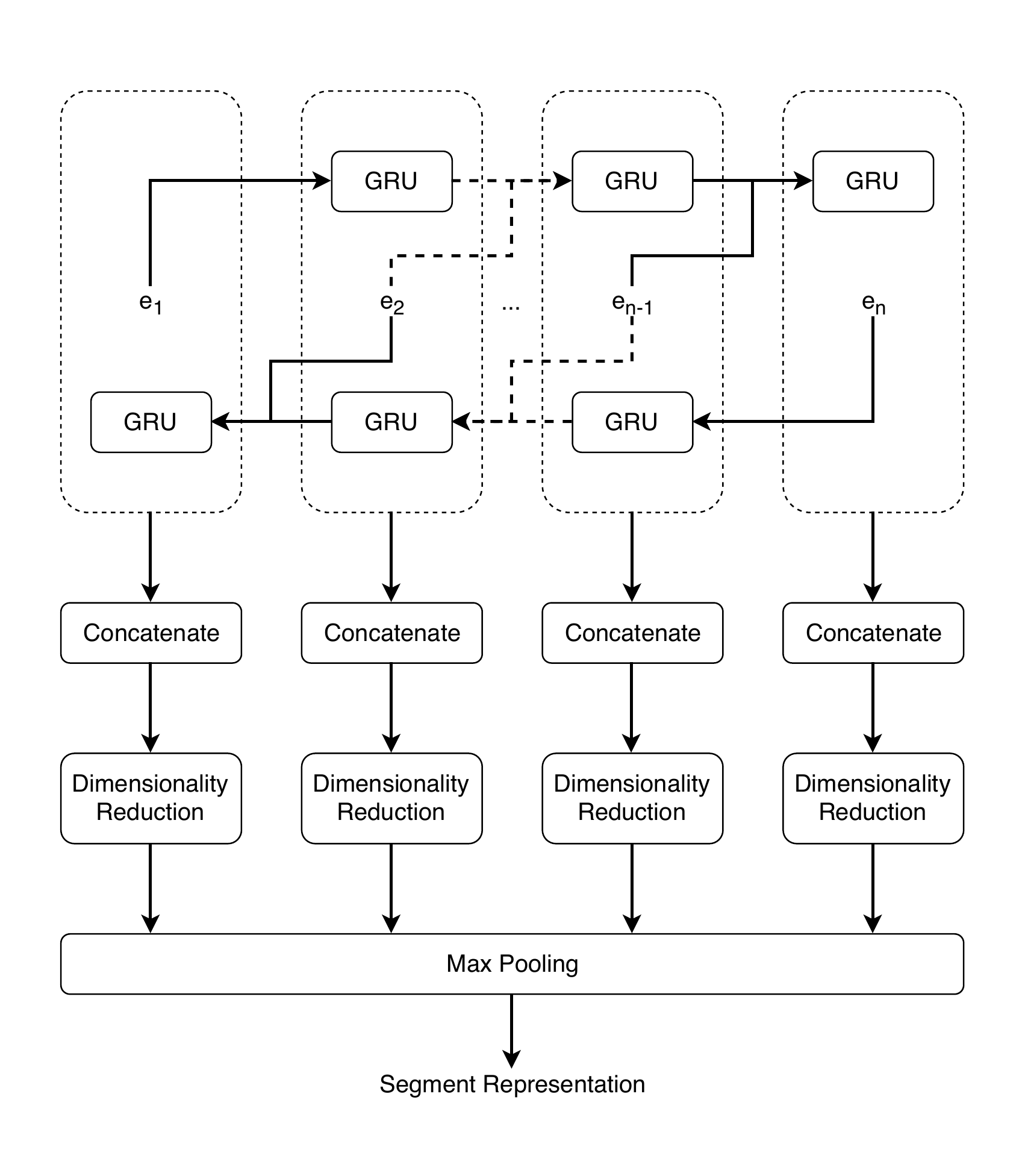}
	\caption{The \ac{RCNN}-based segment representation approach. $e_i$ corresponds to the embedding representation of the $i$-th token.}
	\label{fig:rcnn}
\end{figure}

Of the previous approaches, one focuses on capturing information from relevant sequences of tokens, while the other focuses on capturing information from relevant token patterns independently of where they occur in the segment. The \ac{RCNN}-based approach by \shortciteA{Lai2015} combines some of the advantages of \ac{RNN}- and \ac{CNN}-based segment representation approaches in order to capture both kinds of information. This approach achieved state-of-the-art results on three of the four text classification tasks used to evaluate it, but has not been applied to dialog act recognition yet. Thus, we include it in this study with some adaptations. One of its advantages is that it removes the need for selecting an appropriate context window size for convolution by using a bidirectional recurrent approach which captures context information from all the tokens that appear before and after each token. The embedding representation of each token is then extended by surrounding it with that context information. A linear transformation and the hyperbolic tangent activation are applied to each of those token embeddings to reduce their dimensionality and normalize their representation, respectively. Finally, a max-pooling operation is performed over the sequence of token representations to obtain the segment representation. The process is shown in Figure~\ref{fig:rcnn}.

In the original approach by \shortciteA{Lai2015}, the context information that is appended to the representation of each token is obtained by using basic \acp{RNN} to process the sequences of previous and following tokens. In our experiments, we replace those basic \acp{RNN} with \acp{GRU}, in order to capture relations between distant tokens.

%
%

%
%

\section{Context Information}
\label{sec:context}

Although a dialog act represents the intention behind a set of words, that intention is not constrained to a specific segment and its context provides relevant cues. As described in Section~\ref{sec:related}, previous studies have shown that the most important source of context information for dialog act recognition is the dialog history, with influence decaying with distance~\shortcite{Ribeiro2015,Lee2016,Liu2017}. However, information concerning the speakers and, more specifically, turn-taking has also been proved important~\shortcite{Liu2017}. Thus, in our study, we explore both the surrounding segments and speaker information as sources of context information. 

\subsection{Surrounding Segments}

A dialog is a structured sequence of segments, in which each segment typically depends on both what has been said before and what is expected to be said in future. Thus, the surrounding segments are the most important sources of context information for dialog act recognition. However, although a dialog system may have expectations regarding the future of the dialog, when identifying its conversational partner's intention, it only has access to the dialog history. Thus, we focus on approaches able to capture information from the preceding segments. Still, in order to assess the importance of future information and simulate an annotation scenario, we also perform some experiments using that information.

Considering the preceding segments, we have shown in a previous study~\shortcite{Ribeiro2015} that providing information in the form of segment classifications leads to better results than in the form of words. \shortciteA{Liu2017} have further shown that using a single label per segment is better than using the probability of each class. Furthermore, both studies have shown that using automatic predictions leads to a decrease in performance around two percentage points in comparison to using the manual annotations. Thus, in order to simplify the experiments and obtain an upper bound for the approach, in this study we use the manual annotations during the development phase. During the testing phase we compare the use of the manual annotations and the classifications predicted by the classifier itself, in order to simulate a real scenario. In our previous study, we have used up to five preceding segments and observed that the gain becomes smaller as the number of preceding segments increases, which supports the claim that the closest segments are the most relevant. \shortciteA{Liu2017} stopped at three preceding segments, but noticed a similar pattern. Furthermore, in our study we noticed that the performance tends to stabilize after the third preceding segment. Thus, in this study we explore the use of context information from one and three preceding segments, as well as the entire dialog history.

Although both our previous study and that by \shortciteA{Liu2017} used the classifications of preceding segments as context information, none of them took into account that those segments have a sequential nature and simply flattened the sequence before appending it to the segment representation. However, both studies have shown that each label in that sequence is related to those that precede it. Thus, an approach that captures that information is expected to lead to better performance. To do so, we introduce a recurrent layer that processes the label sequence before appending it to the segment representation. Since the closest segments are expected to have more influence, but there still may be important information in distant segments, we use a \ac{GRU} to process the label sequence. Each element in the sequence output by the recurrent layer consists of information about the sequence of labels up to that element. Thus, the last element of the sequence output by the recurrent layer can be seen as a summary of the sequence of labels of the preceding segments. To assess whether this approach is able to capture information from more distant segments, we compare the use of a summary of the three preceding segments with that of the whole dialog history.

Finally, concerning future information, in terms of representation, the sequence of labels of the following segments can be seen as a mirrored version of the sequence of labels of the preceding segments. Thus, we propose using the same approach based on the processing of the sequence by a recurrent layer. However, in this case, the sequence is processed backwards. To assess the impact of this future information on the task, we explore it both independently of and in combination with preceding segment information.

\subsection{Speaker Information}

Information concerning the speakers is also relevant for dialog act recognition. However, specific information about speaker characteristics may not be available to a dialog system. Still, information about the speaker of each segment, that is, who said what, is always available. Furthermore, intentions may vary if two sequential segments are uttered by the same or different speakers. Thus, turn-taking information is relevant for dialog act recognition. In fact, this has been confirmed in the study by \shortciteA{Liu2017}. In that study, turn-taking information was provided as a flag stating whether the speaker is different from that of the preceding segment. However, there may be relevant information for the task in speaker changes (or the lack thereof) in relation to segments other than the last. Thus, in this study, we explore turn-taking information using an approach similar to that used for preceding segment information. That is, in addition to using a single flag stating whether the speaker changed, we also extend that representation to three preceding segments and the whole turn-taking history. Furthermore, we explore both the flat sequence of flags and the summary generated by the recurrent approach.

%
%

%
%

\section{Experimental Setup}
\label{sec:setup}

In this section, we describe our experimental setup, starting with the datasets used in our experiments. Then, we define a generic architecture that sets a common ground for result comparison. We also describe our evaluation approach. Furthermore, we provide implementation details to allow future reproduction of all of our experiments.

%
%

\subsection{Datasets}
\label{ssec:datasets}

We selected two datasets for our experiments {--} the \acf{SwDA} and the \acf{MRDA} {--}, which, to the best of our knowledge, are the largest corpora with generic dialog act annotations. Considering our focus on the application of \ac{DNN}-based approaches, the size of the datasets is an important factor to avoid overfitting. Furthermore, both corpora feature dialogs with varying domain, which is important in order to avoid drawing domain-dependent conclusions. Finally, both datasets have been previously explored for the task. Thus, the results of previous studies provide a performance baseline.

\subsubsection{Switchboard Dialog Act Corpus}

Switchboard~\shortcite{Godfrey1992} is a corpus consisting of 2,430 telephone conversations among 543 American English speakers (302 male and 241 female). Each pair of speakers was automatically attributed a topic for discussion, from 70 different ones. Furthermore, speaker pairing and topic attribution were constrained so that no two speakers would be paired with each other more than once and no one spoke more than once on a given topic.

\begin{figure}[ht]
    \begin{framed}
    \begin{dialogue}
        \speak{Spk 1} Well, have you ever served on a jury? \hfill \textbf{Yes-No-Question}
		\speak{Spk 2} No, \hfill \textbf{No Answer}
        \speak{Spk 2} I've not. \hfill \textbf{Statement-Non-Opinion}
        \speak{Spk 2} I've been called, \hfill \textbf{Statement-Non-Opinion}
        \speak{Spk 2} but I had to beg off from the duty. \hfill \textbf{Statement-Non-Opinion}
        \speak{Spk 2} And you? \hfill \textbf{Yes-No-Question} 
        \speak{Spk 1} Well, I was called \hfill \textbf{Affirmative Non-Yes Answer}
        \speak{Spk 1} and then I was not chosen. \hfill \textbf{Statement-Non-Opinion}
        \speak{Spk 2} Well, I was originally chosen primarily, \\ I think, because I was a young fellow \hfill \textbf{Statement-Opinion}
        \speak{Spk 2} and they tend to view the younger fellows \\ as more likely to hand down a guilty verdict. \hfill \textbf{Statement-Opinion}
        \speak{Spk 2} I don't know why. \hfill \textbf{Statement-Non-Opinion}
        \speak{Spk 2} Something I picked up in a psychology class \\ some time ago. \hfill \textbf{Statement-Non-Opinion}
        \speak{Spk 1} Oh, really? \hfill \textbf{Backchannel-Question}
    \end{dialogue}
    \end{framed}
    \caption{An excerpt of a \acf{SwDA} transcription with the corresponding dialog act annotations. For readability, we removed disfluency information.}
    \label{fig:swdadialog}
\end{figure}

The \acf{SwDA}~\shortcite{Jurafsky1997} is a subset of this corpus, consisting of 1,155 manually transcribed conversations, containing 223,606 segments. An excerpt of a transcription is shown in Figure~\ref{fig:swdadialog}. The corpus was annotated for dialog acts using the SWBD-DAMSL tag set, which was structured so that the annotators were able to label the conversations from transcriptions alone. Including combinations, there were 220 unique tags in the annotated segments. However, in order to obtain a higher inter-annotator agreement and higher example frequencies per class, a less fine-grained set of 44 tags was devised. As shown in Table~\ref{tab:swdadistr}, the class distribution is highly unbalanced, with the three most frequent classes {--} \textit{Statement-Opinion}, \textit{Acknowledgement}, and \textit{Statement-Non-Opinion} {--} covering 68\% of the corpus. The set can be reduced to 43 or 42 categories~\shortcite{Stolcke2000,Rotaru2002,Gamback2011}, if the \textit{Abandoned} and \textit{Uninterpretable} categories are merged, and depending on how the \textit{Segment} category, used when the current segment is the continuation of the previous one by the same speaker, is treated. By analyzing the data, we came to the conclusion that merging segments labeled as \textit{Segment} with the previous segment by the same speaker is an adequate approach, since some of the attributed labels only make sense when the segments are merged. Also, it makes sense to merge the \textit{Abandoned} and \textit{Uninterpretable} categories, because both represent disruptions in the dialog flow, which interfere with the typical dialog act sequence. There is also a 41-category variant of the tag set~\shortcite{Webb2010}, which merges the \textit{Statement-Opinion} and \textit{Statement-Non-Opinion} categories, making this the most frequent class, covering 49\% of the corpus. \shortciteA{Jurafsky1997} report an average pairwise Kappa~\shortcite{Carletta1996} of 0.80, while \shortciteA{Stolcke2000} refer to an inter-annotator agreement of 84\%, which is the average pairwise percent agreement when considering 42 categories.

\begin{table}[ht]
\begin{center}
    \begin{tabular}{l r r | l r r}
        \toprule
        \textbf{Label} & \textbf{Count} & \textbf{\%} & \textbf{Label} & \textbf{Count} & \textbf{\%}  \tabularnewline
        \midrule
        Statement-Non-Opinion       & 72,824 &  36  & Collaborative Completion &   699 &          0.4  \tabularnewline
        Acknowledgement             & 37,096 &  19  & Repeat-Phrase            &   660 &          0.3  \tabularnewline
        Statement-Opinion           & 25,197 &  13  & Open-Question            &   632 &          0.3  \tabularnewline
        Agreement                   & 10,820 &   5  & Rhetorical-Question      &   557 &          0.3  \tabularnewline
        Abandoned                   & 10,569 &   5  & Hold                     &   540 &          0.2  \tabularnewline
        Appreciation                &  4,663 &   2  & Reject                   &   338 &          0.2  \tabularnewline
        Yes-No-Question             &  4,624 &   2  & Negative Non-No Answer   &   292 &          0.1  \tabularnewline
        Non-Verbal                  &  3,548 &   2  & Non-understanding        &   288 &          0.1  \tabularnewline
        Yes Answer                  &  2,934 &   1  & Other Answer             &   279 &          0.1  \tabularnewline
        Conventional Closing        &  2,486 &   1  & Conventional Opening     &   220 &          0.1  \tabularnewline
        Uninterpretable             &  2,158 &   1  & Or-Clause                &   207 &          0.1  \tabularnewline
        Wh-Question                 &  1,911 &   1  & Dispreferred Answers     &   205 &          0.1  \tabularnewline
        No Answer                   &  1,340 &   1  & 3rd-Party-Talk           &   115 &          0.1  \tabularnewline
        Response Acknowledgement    &  1,277 &   1  & Offers / Options         &   109 &          0.1  \tabularnewline
        Hedge                       &  1,182 &   1  & Self-talk                &   102 &          0.1  \tabularnewline
        Declarative Yes-No-Question &  1,174 &   1  & Downplayer               &   100 &          0.1  \tabularnewline
        Other                       &  1,074 &   1  & Maybe                    &    98 & \textless0.1  \tabularnewline
        Backchannel-Question        &  1,019 &   1  & Tag-Question             &    93 & \textless0.1  \tabularnewline
        Quotation                   &    934 & 0.5  & Declarative Wh-Question  &    80 & \textless0.1  \tabularnewline
        Summarization               &    919 & 0.5  & Apology                  &    76 & \textless0.1  \tabularnewline
        Affirmative Non-Yes Answer  &    836 & 0.4  & Thanking                 &    67 & \textless0.1  \tabularnewline
        Action Directive            &    719 & 0.4  &                          &       &              \tabularnewline
        \bottomrule
    \end{tabular}
\end{center}
\caption{Label distribution in the \acf{SwDA}~\shortcite{Jurafsky1997}.}
\label{tab:swdadistr}
\end{table}

\subsubsection{ICSI Meeting Recorder Dialog Act Corpus}

The ICSI Meeting Corpus~\shortcite{Janin2003} consists of 75 meetings, each lasting around an hour. Of these meetings, 29 are of the project itself, 23 are of a research group
focused on robustness in automatic speech recognition, 15 involve a group discussing natural language
processing and neural theories of language, and 8 are miscellaneous meetings. There are 53 speakers in the corpus (40 male and 13 female), out of which 28 are native English speakers. Contrarily to Switchboard, which features two speakers per dialog, the average number of speakers per meeting is six.

\begin{figure}[ht]
    \begin{framed}
    \begin{dialogue}
        \speak{Spk 1} Ok, do you go around the room and do names or anything? \hfill \textbf{Question}
		\speak{Spk 2} I think that \hfill \textbf{Disruption}
        \speak{Spk 3} It's a good idea. \hfill \textbf{Statement}
        \speak{Spk 2} usually we've done that. \hfill \textbf{Statement}
        \speak{Spk 2} And also we've done digits as well, \hfill \textbf{Statement}
        \speak{Spk 2} but I forgot to print any out. \hfill \textbf{Statement} 
        \speak{Spk 2} So... \hfill \textbf{Filler}
        \speak{Spk 2} Besides, with this big a group \hfill \textbf{Statement}
        \speak{Spk 2} it would take too much time. \hfill \textbf{Disruption}
        \speak{Spk 4} No, it'd be even better with this big \hfill \textbf{Disruption}
        \speak{Spk 5} You can write them on the board if you want. \hfill \textbf{Statement}
    \end{dialogue}
    \end{framed}
    \caption{An excerpt of an \acf{MRDA} transcription with the corresponding dialog act annotations. We have curated the transcription for readability by removing disfluency information and adding casing information.}
    \label{fig:mrdadialog}
\end{figure}

The \acf{MRDA}~\shortcite{Shriberg2004} is a set of annotations that augment the word transcriptions of the ICSI Meeting Corpus with discourse-level segmentation, dialog act information, and adjacency pair information. An excerpt of one of the transcriptions is shown in Figure~\ref{fig:mrdadialog}. An adapted version of the SWDA-DAMSL tag set was used to annotate the corpus. It features 11 general tags, one of which must be attributed to each segment, and 39 specific tags, which specialize the general tags. However, the studies on automatic dialog act recognition on this corpus, starting with that by \shortciteA{Ang2005}, reduced the tag set to five generic classes {--} \textit{Statement}, \textit{Disruption}, \textit{Backchannel}, \textit{Filler}, and \textit{Question}. There is an additional tag for unlabeled segments. However, we discarded those segments in our experiments. The annotations of the remaining 106,360 annotated segments follow the distribution shown in Table~\ref{tab:mrdadistr}. We can see the distribution is highly unbalanced, with segments labeled as \textit{Statement} covering 59\% of the corpus. In terms of inter-annotator agreement, \shortciteA{Shriberg2004} report an average pairwise Kappa of 0.80 when considering these five classes.

\begin{table}[ht]
\begin{center}
    \begin{tabular}{l r r}
        \toprule
        \textbf{Label} & \textbf{Count} & \textbf{\%} \tabularnewline
        \midrule
        Statement   & 62,608 &  59 \tabularnewline
        Disruption  & 14,978 &  14 \tabularnewline
        Backchannel & 14,275 &  13 \tabularnewline
        Filler      &  7,701 &   7 \tabularnewline
        Question    &  6,798 &   6 \tabularnewline
        \bottomrule
    \end{tabular}
\end{center}
\caption{Label distribution in the 5-class version of the \acf{MRDA}~\shortcite{Shriberg2004}.}
\label{tab:mrdadistr}
\end{table}

%
%

%
%

\subsection{Generic Architecture for Dialog Act Recognition}
\label{ssec:architecture}

\begin{figure}[ht]
	\centering
	\includegraphics[scale=0.6]{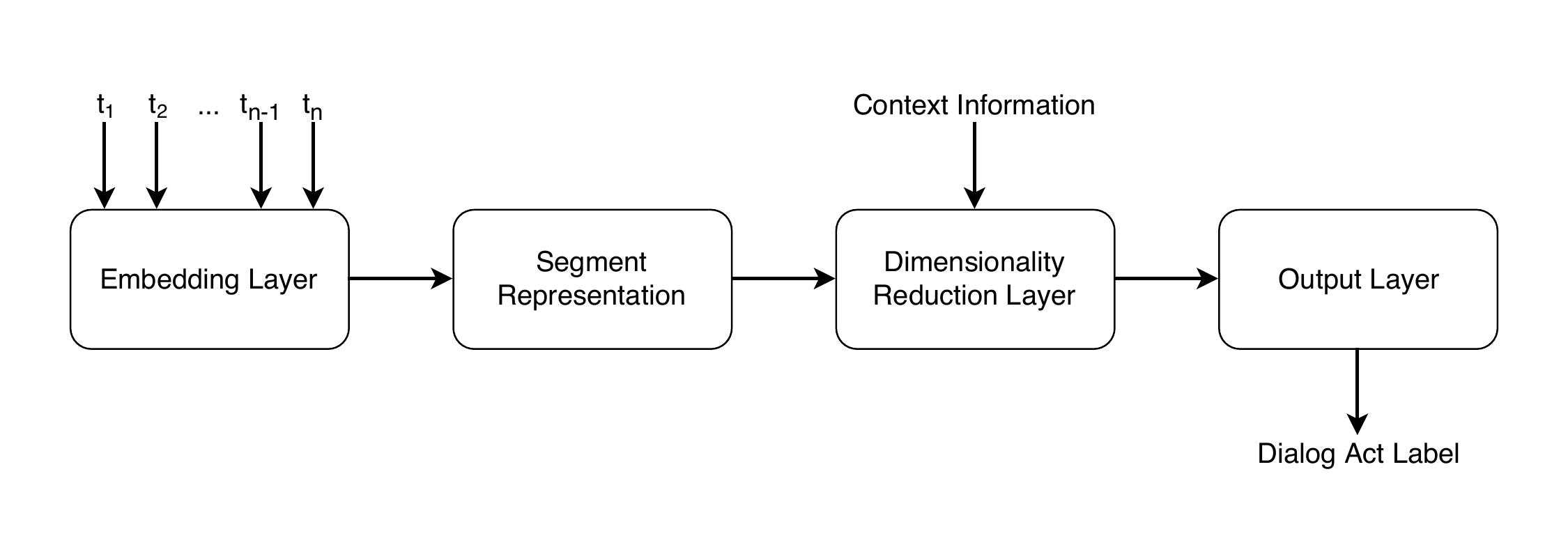}
	\caption{The generic architecture used throughout our experiments. The input, $t_1, t_2, ..., t_n$, consists of the sequence of tokens in a segment. Those tokens are then mapped into an embedding space in the embedding layer. A representation of the segment is generated by combining the embedding representations of its tokens. Additional contextual information can be appended to the generated representation before its dimensionality is reduced for normalization. Finally, the label probability distribution for that reduced representation is computed to identify the label with highest probability.}
	\label{fig:arch}
\end{figure}

\begin{figure}[p]
	\centering
	\includegraphics[scale=0.6]{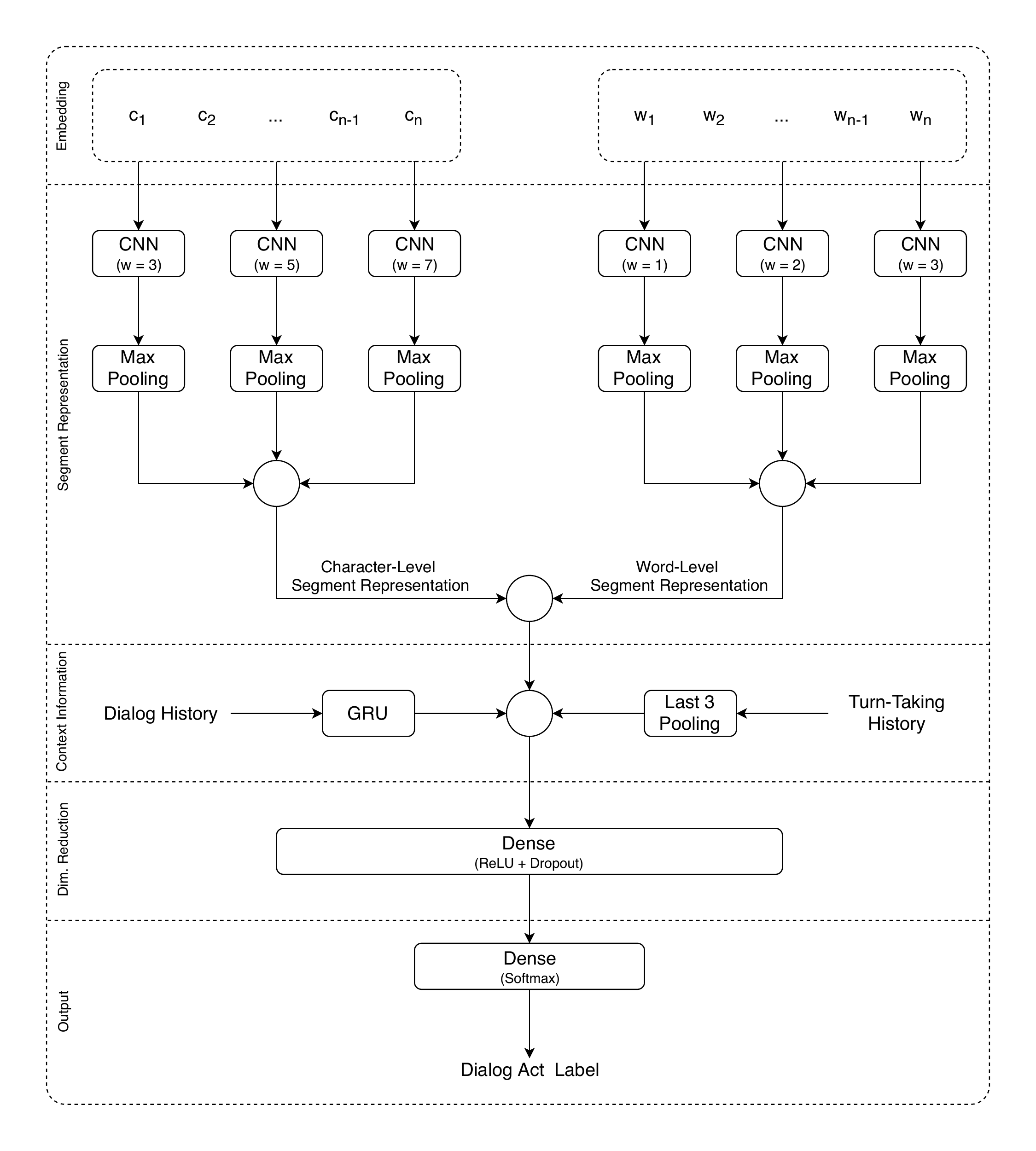}
	\caption{The expanded architecture of the approach with best performance in our experiments. The circles correspond to concatenation operations. The main inputs, $c_1, c_2, ..., c_n$ and $w_1, w_2, ..., w_n$, correspond to the embedding representations of the characters and words in the segment, respectively. While the first are trained with the network, the latter are provided by the BERT model. Both the character- and word-level segment representations are generated using a \ac{CNN}-based approach, but with different context windows $w$. Context information from the dialog history is provided as a summary of the classifications of the surrounding segments. Turn-taking information from the three preceding segments is provided in the form of a flat sequence of flags.}
	\label{fig:fullarch}
\end{figure}

In order to set a common ground for result comparison, we use the generic architecture shown in Figure~\ref{fig:arch}, which is based on those of the top performing approaches presented in Section~\ref{sec:related}. Below, we describe each of its components and their initial state. The experiments are then built incrementally to select the best approach for each component, among those described in Sections~\ref{sec:token}, \ref{sec:segment}, and \ref{sec:context}, respectively. The expanded architecture of our best approach is shown in Figure~\ref{fig:fullarch}.

\subsubsection{Embedding Layer}

The input of our system is a tokenized segment, which is passed to an embedding layer that maps each token into its representation in an embedding space. As a baseline, we use the approach described in Section~\ref{sssec:random}. That is, we perform tokenization at the word level and do not use pre-trained embeddings. Thus, the weights of the embedding layer are initialized randomly and updated along with the rest of the network. This means that the generated word embeddings are fit to the training corpus and influenced by the task. 

\subsubsection{Segment Representation}

The segment representation step processes and combines the token embeddings to generate a vectorial representation of the segment. This is the step that introduces higher variability in the network and in which the main differences between previous dialog act recognition approaches occur. Initially, to keep the consistence with the character-level approach, we use the \ac{CNN}-based approach for segment representation described in Section~\ref{ssec:cnn}.

In addition to the approaches described in Section~\ref{sec:segment}, we perform an experiment which reduces the representation of a segment to the area in the embedding space that contains it by performing a max-pooling operation over the embeddings of its tokens. This approach serves as a baseline which allows us to assess the importance of a complex segment representation approach for the task.

In experiments which combine tokenization at multiple levels, both the Embedding and Segment Representation layers are replicated for each level. The segment representation is then given by the concatenation of the representations generated by each approach, as exemplified in Figure~\ref{fig:fullarch} for the combination of word and character levels.

\subsubsection{Context Information}

\shortciteA{Liu2017} have shown that approaches that concatenate context information directly to the segment representation perform better than discourse models. Thus, we use the same approach to provide context information to the network. In the approaches that generate a summary of the dialog history, the sequence of classifications is passed through a recurrent layer before the concatenation to the segment representation. This is exemplified in Figure~\ref{fig:fullarch}, in which the whole dialog history is passed through a \ac{GRU} to generate its summary. The same occurs when using approaches that generate a summary of the turn-taking history.

In order to reduce the number of variables, context information is not taken into account in the experiments to select the best token and segment representation approaches. Thus, the impact of context information is assessed in specialized experiments, with fixed token and segment representation approaches.

\subsubsection{Dimensionality Reduction Layer}

In order to avoid result differences caused by using representations with different dimensionality, the network includes a dimensionality reduction layer. This is a dense layer which maps the segment representations, with or without concatenated context information, into a reduced space with fixed dimensionality. Furthermore, this layer reduces the probability of overfitting to the training data by applying dropout during the training phase.

\subsubsection{Output Layer}

Finally, the output layer maps the reduced representation into a dialog act label. To do so, we use a dense layer with a number of units equal to the number of labels, in which each unit represents a possible category. To calculate the probability distribution over the possible labels this layer uses the softmax activation. The class with highest probability is then selected for the segment. Since we are dealing with a multiclass classification problem, we use the categorical cross entropy loss function. 

%
%
%
%

\subsection{Implementation Details}
\label{ssec:implementation}

To implement the networks we used Keras~\shortcite{Chollet2015} with the TensorFlow~\shortcite{Abadi2015} backend. For performance reasons, we used the Adam optimizer~\shortcite{Kingma2015} to update the weights during the training phase. The results reported in the next section were achieved when using mini-batches with size 512 and early stopping with ten epochs of patience. That is, the training phase stopped after ten epochs without accuracy improvement on the validation set. Implementation details of each layer are described below.

Starting with the experiments using word-level tokenization and, more specifically, pre-trained word embeddings, we kept the original dimensionality of each set of embeddings. For the Word2Vec model, we used the same embeddings as \shortciteA{Khanpour2016}, which were trained on Wikipedia data. For FastText, we used the embeddings with subword information provided by \shortciteA{Mikolov2018}, which were trained on Wikipedia, the UMBC webbase corpus~\shortcite{Han2013}, and the statmt.org news dataset~\shortcite{Tiedemann2012}. For the dependency-based embeddings, we used the pre-trained set provided by \shortciteA{Levy2014}, which was trained on Wikipedia data. All of the previous have dimensionality 300. However, in the experiments with randomly initialized embeddings, we used dimensionality 200 for consistency with the study by \shortciteA{Liu2017}. To obtain the contextualized embeddings, we used the ELMo model provided by the AllenNLP package~\shortcite{Gardner2017} and the large uncased BERT model provided by its authors~\shortcite{Devlin2018}, both of which generate embeddings with dimensionality 1,024.

To obtain the lemmas and the \ac{POS} tags of each word to use in the experiments with functional level tokenization, we used the spaCy parser~\shortcite{Honnibal2017}. In both this scenario and the one using character-level tokenization, the dimensionality of the embedding space is equivalent to that necessary for a one-hot encoding of each token.

The adaptations to the \ac{RNN}- and \ac{CNN}-based approaches described in Sections~\ref{ssec:rnn} and~\ref{ssec:cnn}, respectively, did not lead to significant improvement in practice. Thus, the results reported in the next section use the original configuration of both approaches. That is, a stack of 10 \acp{LSTM} with a number of neurons equal to the dimensionality of the embeddings for the \ac{RNN}-based approach and a set of three parallel \acp{CNN} with 100 filters each for the \ac{CNN}-based approach. We used context window sizes of one, two, and three for the word- and functional-level approaches and three, five, and seven for the character-based approach. For the \ac{RCNN}-based approach, we used contextual \acp{GRU} with 200 neurons each and a dimensionality reduction layer with 200 neurons as well. That is, the embedding representation of each token is decorated with 200-dimensional summaries of the preceding and future tokens and, then, the dimensionality of the decorated representation is reduced to 200.

When providing context information from the surrounding segments as a flattened sequence, we encoded each element of the sequence as a one-hot vector representing the classification. When the sequence is summarized, the \ac{GRU} had a number of neurons equal to the dimensionality of those vectors.

Finally, the dimensionality reduction layer is implemented as a \ac{ReLU} with 100 neurons and 50\% dropout probability.

%
%
%
%

\subsection{Evaluation Approach}
\label{ssec:evaluation}

In all of the studies presented in Section~\ref{sec:related}, the performance on the task was evaluated using accuracy as metric. Thus, for consistency and to allow comparison with those studies, we also use accuracy to evaluate the performance of our approaches. In order to attenuate the influence of random initialization and the non-determinism of some operations ran on \ac{GPU}, the results presented in the remainder of the article refer to the average ($\mu$) and standard deviation ($\sigma$) of the accuracy results obtained by training and testing the networks over ten runs. Furthermore, for readability, we present the results in percentage form.

We stop training the networks after ten epochs without improvement on the validation set. Furthermore, our experiments are built incrementally according to the results on the validation set and we only report test set results for the final approach, in comparison to those of previous studies. Thus, it is important to define the partitions of the two datasets used in this study. In both cases, we use the standard partitions defined in previous studies. For \ac{SwDA}, \shortciteA{Stolcke2000} describe a data partition of the corpus into a training set of 1,115 conversations, a test set of 19 conversations, and a future use set of 21 conversations. In our experiments, we use the latter as a validation set. \ac{MRDA} is shared with a standard data partition with 51 dialogs for training, 11 for development/validation, and 11 for testing.

Since there are variations in the multiple runs for each approach, it is difficult to assess whether the performance difference between two approaches is statistically significant using tests based on confusion matrices. Furthermore, performing a simple binomial test on the average accuracy of the approaches is not fair, since, given the size differences between the training, validation, and test sets used in our experiments, the exact same difference may be classified as significant in some cases and not in others. Thus, in the discussion in the next section, we use the word significant to state that the sum of the average and standard deviation accuracy of the approach with worst performance does not surpass the subtraction of the standard deviation from the average accuracy of the approach with best performance. That is, we consider the difference to be significant if $ \mu_w + \sigma_w < \mu_b - \sigma_b $, where $w$ and $b$ stand for the approaches with worst and best performance, respectively.  

%
%

%
%
%
%

\section{Results \& Discussion}
\label{sec:results}

Since our experiments were built incrementally, we start by presenting the results achieved on the validation sets using the different token representation approaches. Then, we build on the best result and explore the different segment representation approaches. Similarly, after that, we present the results achieved using context information from different sources. Then, we report the results achieved by the best approach, that is, the combination of the best representation approaches for each step, on the test sets. Finally, we compare the results achieved using our approach with those of previous studies.

\subsection{Results Depending on Token Representation}

Table~\ref{tab:restoken} presents the results achieved using different token representation approaches. Remember that in these experiments we used the \ac{CNN}-based segment representation approach and did not consider context information. In the first block we can see the results achieved using word-level approaches. In comparison to the random initialization, the highest improvement achieved using pre-trained uncontextualized embeddings was of 1.02 percentage points on \ac{SwDA} and 1.42 on \ac{MRDA}. These results were achieved using the dependency-based embeddings, which confirms that the segment structure information included in the representation is relevant for the task. A more surprising result was the fact that FastText embeddings were not able to outperform Word2Vec embeddings on \ac{SwDA}. However, this is explainable by the simple use of the pre-trained embeddings for each word, without replicating the embedding approach. This way, only the words present in the set of pre-trained embeddings are considered and, consequently, the ability to deal with \acp{OOV}, which is an advantage of FastText in comparison to Word2Vec, is lost.

\begin{table} [ht]
\begin{center}
    \begin{tabular}{l c c c c}
        \toprule
                                &  \multicolumn{2}{c}{\textbf{SwDA}} &  \multicolumn{2}{c}{\textbf{MRDA}} \tabularnewline
        \textbf{Approach}       & \textbf{$\mu$} & \textbf{$\sigma$} & \textbf{$\mu$} & \textbf{$\sigma$} \tabularnewline
        \midrule      
        Random Initialization   &          76.02 &              0.18 &          76.94 &              0.23 \tabularnewline
        Word2Vec                &          76.58 &              0.26 &          77.67 &              0.04 \tabularnewline
        FastText                &          76.38 &              0.25 &          78.10 &              0.14 \tabularnewline
        Dependency-Based        &          77.04 &              0.20 &          78.36 &              0.07 \tabularnewline
        \hdashline
        ELMo                    &          77.85 &              0.24 &          80.67 &              0.14 \tabularnewline
        BERT                    &          79.19 &              0.16 &          88.62 &              0.03 \tabularnewline
        \midrule                      
        Character-Level         &          76.47 &              0.29 &          88.66 &              0.02 \tabularnewline
        \midrule                      
        Coarse-Grained POS      &          66.91 &              0.33 &          80.57 &              0.12 \tabularnewline
        Fine-Grained POS        &          69.47 &              0.15 &          82.12 &              0.05 \tabularnewline
        Lemma                   &          74.46 &              0.24 &          77.05 &              0.13 \tabularnewline
        \midrule                      
        BERT + Char-Level       &          79.34 &              0.10 &          88.69 &              0.02 \tabularnewline
        BERT + Fine-Grained POS &          79.09 &              0.30 &          88.62 &              0.05 \tabularnewline
        \bottomrule     
    \end{tabular}
\end{center}
\caption{Accuracy (\%) results using different token representation approaches. The first block refers to word-level approaches, the second to the character-level approach, the third to functional-level approaches, and the last to the combination of multiple levels. The separation (dashed line) in the block for word-level approaches distinguishes between uncontextualized and contextualized word representations.}
\label{tab:restoken}
\end{table}

The most relevant improvements at the word-level were achieved when using contextualized word representations, especially those generated by BERT. The improvement in relation to when using dependency-based embeddings was of 2.15 percentage points on \ac{SwDA} and 10.26 on \ac{MRDA}. When using ELMo embeddings, this improvement is reduced to 0.81 and 2.31 percentage points, respectively. The higher improvement on the \ac{MRDA} corpus is explained by the reduced number of classes, which are easily distinguishable by the context in which some words appear. Still, the nearly eight point difference between ELMo and BERT embeddings on this corpus suggests that the attention-based architecture focuses on aspects that are more relevant for the distinction of more abstract dialog acts. However, we leave the study of this phenomenom for future work.

According to our previous studies~\shortcite{Ribeiro2018,Ribeiro2019}, using character-level tokenization on \ac{SwDA} leads to results similar to those achieved using pretrained Word2Vec embeddings. On \ac{MRDA}, the character-level approach surpassed every word-level approach. There was an improvement of over 10 percentage points in relation to when using Word2Vec embeddings, but only a slight improvement in relation to when using BERT embeddings. The higher performance of the character-level approach on \ac{MRDA} is due to the reduced number of classes, which are not strongly connected to the presence of specific words, but rather to the function of those words, for which the morphological aspects captured by the character-level approach are able to provide cues. In fact, when considering domain-independent dialog acts, our previous study~\shortcite{Ribeiro2019} had already revealed a higher performance of the character-level approach on the DIHANA~\shortcite{Benedi2006} and VERBMOBIL~\shortcite{Alexandersson1998} corpora, which have less classes than \ac{SwDA}. 

The third block in Table~\ref{tab:restoken} shows the results achieved using functional-level approaches. We can see that, in this case, the performance is also highly dependent on the granularity of the dialog acts that is being considered, especially when using tokenization at the \ac{POS} tag level. While on \ac{SwDA} the results were much lower than those achieved by the word- and character-level approaches, on \ac{MRDA} they surpassed those of every word-level approach except BERT. This confirms that the five classes used in \ac{MRDA} are highly related to the function of the words and the structure of the segment, and not to the presence of specific words. On the other hand, in \ac{SwDA} there are more specific classes, which require word-specific information to be identified. However, in both cases, the set of fine-grained \ac{POS} tags containing morphological information performed better than the coarse-grained set which only states the word type. This is in line with the findings at the character-level, which have shown the importance of morphological information for the task. As future work, it would be interesting to assess whether there are specific word-classes which can be abstracted or if the specific words required to identify some \ac{SwDA} dialog acts are scattered over all word types.

By looking at the results achieved using tokenization at the lemma level we can see that, on \ac{SwDA}, the performance is 1.56 percentage points lower than when using word-level tokenization with a random initialization. This suggests that most intentions can be transmitted using a simplified version of the language consisting of just the lemmas and that the affixes are only used to transmit more specific intentions. This is confirmed by the results on \ac{MRDA}, in which there is no significant difference between the two approaches.

Finally, the last block in Table~\ref{tab:restoken} shows the results achieved using the combination of approaches at multiple levels. We can see that the best results on both corpora were achieved by the combination of the character-level approach with the best word-level approach. This is in line with the conclusions of our previous studies~\shortcite{Ribeiro2018,Ribeiro2019}, which have shown that the word- and character-level are able to capture complementary information. Still, it is interesting to observe that this holds even when using contextualized embeddings, which are able to capture more information concerning the function of the word than the uncontextualized embeddings. On the other hand, the combination of the word- and functional-level approaches was not able to improve the results of the best word-level approach on its own. Thus, we did not explore the combination of the three levels.

\subsection{Results Depending on Segment Representation}

Table~\ref{tab:ressegment} presents the results achieved using different segment representation approaches. Since the experiments were built incrementally, the results were obtained using the combination of the word- and character-level token representation approaches, with BERT embeddings at the word level. However, for the character-level approach, the segment representation is always generated using the \ac{CNN}-based approach. Thus, the results in this section refer only to changes in the approach used to generate the segment representation for the word-level approach. 

\begin{table} [ht]
\begin{center}
    \begin{tabular}{l c c c c}
        \toprule
                           &  \multicolumn{2}{c}{\textbf{SwDA}} &  \multicolumn{2}{c}{\textbf{MRDA}} \tabularnewline
        \textbf{Approach}  & \textbf{$\mu$} & \textbf{$\sigma$} & \textbf{$\mu$} & \textbf{$\sigma$} \tabularnewline
        \midrule      
        Max Pooling        &          78.58 &              0.15 &          88.67 &              0.02 \tabularnewline
        Parallel \acp{CNN} &          79.34 &              0.10 &          88.69 &              0.02 \tabularnewline
        \ac{LSTM} Stack    &          79.33 &              0.27 &          88.70 &              0.02 \tabularnewline
        \ac{RCNN}          &          79.35 &              0.09 &          88.70 &              0.02 \tabularnewline
        \bottomrule     
    \end{tabular}
\end{center}
\caption{Accuracy (\%) results using different segment representation approaches.}
\label{tab:ressegment}
\end{table}

We can see that there is no significant difference in terms of results between the \ac{CNN}- and \ac{RNN}-based approaches. However, the training of the \ac{LSTM} stack used by the \ac{RNN}-based approach is much more resource consuming, taking around 9 times more memory and 31 times longer per epoch. Furthermore, the \ac{CNN}-based approach is also faster during the inference phase. Thus, it is more appropriate in the context of a dialog system that must predict the intention of its conversational partners. 

Although the \ac{RCNN}-based approach was designed to combine the advantages of the \ac{CNN}- and \ac{RNN}-based approaches, it was not able to significantly improve the results. However, this is due to the use of BERT embeddings, which already include contextual information from all the tokens in the sentence in the representation of each word. Thus, the application of the \ac{RCNN} is redundant. In fact, we can see that by considering a simple baseline in which the segment representation was given by the result of a max pooling operation applied to the representation of all tokens in the segment, the obtained results were less than one percentage point below the best approach on \ac{SwDA} and in line with the results of the more complex approaches on \ac{MRDA}. The max pooling operation applied to the representation of the tokens defines the area in the embedding space in which the segment is contained. The fact that it achieves results close to the more complex segment representation approaches means that the representations of the tokens define an area in the embedding space which contains enough information to disambiguate intention. On the other hand, since uncontextualized embeddings do not provide information concerning the context in which the word is used, when the max pooling baseline is applied to them, the obtained results are worse than those achieved by the \ac{CNN}- and \ac{RNN}-based approaches. Furthermore, in that case, the \ac{RCNN}-based approach significantly improves the performance on the task. 

Overall, since all the approaches performed similarly when applied to token representations in the form of BERT embeddings, we decided to use the \ac{CNN}-based approach in the remainder of the experiments, since it consumes less resources than the \ac{RNN}- and \ac{RCNN}-based approaches.

\subsection{Results Depending on Context Information}

As discussed in Section~\ref{sec:context}, in our experiments, we explored two sources of context information: the intention of the surrounding segments in the form of their dialog act classifications and the turn-taking history. We started by assessing the influence of the information provided by each source independently, using different approaches, and then combined information from the two sources to achieve the best performance on the task.

Table~\ref{tab:ressurrounding} shows the results achieved when including context information in the form of the classifications of the surrounding segments. By providing information from the first preceding segment, the performance improved by 2.38 percentage points on \ac{SwDA} and 0.26 on \ac{MRDA}. By considering two additional segments in a flattened sequence of classifications, there was an additional improvement of 0.66 percentage points on \ac{SwDA} and 0.21 on \ac{MRDA}. This influence decrease with the distance is consistent with the findings of previous studies~\shortcite{Ribeiro2015,Liu2017}. However, on \ac{SwDA}, the improvement achieved by using context information from three preceding segments was less than the reported on those studies. This is due to the use of BERT embeddings and the character-level approach, which enables the disambiguation of certain classes that could only be disambiguated using context in those studies. Furthermore, the influence of context information from these segments is less pronounced on \ac{MRDA} since the classes are easier to predict independently.

\begin{table} [ht]
\begin{center}
    \begin{tabular}{l c c c c}
        \toprule
                              &  \multicolumn{2}{c}{\textbf{SwDA}} &  \multicolumn{2}{c}{\textbf{MRDA}} \tabularnewline
        \textbf{Approach}     & \textbf{$\mu$} & \textbf{$\sigma$} & \textbf{$\mu$} & \textbf{$\sigma$} \tabularnewline
        \midrule      
        Without Context       &          79.34 &              0.10 &          88.69 &              0.02 \tabularnewline
        \midrule
        1 Preceding           &          81.72 &              0.40 &          88.95 &              0.03 \tabularnewline
        3 Preceding Flat      &          82.38 &              0.21 &          89.16 &              0.04 \tabularnewline
        Whole History Flat    &          81.88 &              0.25 &          79.01 &              0.04 \tabularnewline
        3 Preceding Summary   &          82.49 &              0.22 &          89.16 &              0.03 \tabularnewline
        Whole History Summary &          82.76 &              0.13 &          89.19 &              0.01 \tabularnewline
        \midrule
        Future Summary        &          81.53 &              0.31 &          88.86 &              0.04 \tabularnewline
        \midrule
        Preceding + Future    &          84.15 &              0.27 &          89.38 &              0.04 \tabularnewline
        \bottomrule     
    \end{tabular}
\end{center}
\caption{Accuracy (\%) results using context information from the surrounding segments.}
\label{tab:ressurrounding}
\end{table}

In Table~\ref{tab:ressurrounding} we can also see that providing information from the whole dialog history, in the form of a flattened sequence, impairs the performance, since, given its implementation, it introduces a large amount of noise in the form of padding. However, while on \ac{MRDA} the results are nearly 10 percentage points below the scenario without context information, on \ac{SwDA} the results are still above the scenario with context information from one preceding segment. On the other hand, if the sequence is summarized using the recurrent approach, at least part of the noise is discarded and additional information can be extracted, as revealed by the improvement on \ac{SwDA} in relation to the scenario with information from three preceding segments as a flat sequence. Finally, the results achieved using a summary of the classifications of just three preceding segments were in line or above those of the achieved using the flat approach, but never above those achieved using the summary of the whole dialog history. This confirms that there is relevant information on more distant segments, which the summary is able to capture.

By summarizing the classifications of future segments in the same manner, the achieved results were 2.19 percentage points above those of the scenario without context information on \ac{SwDA} and 0.17 on \ac{MRDA}. This shows that, as expected, the intention of a segment also depends on the speaker's expectations of the future of the dialog. However, the observed improvements were below the 3.42 and 0.5 achieved when using information from the preceding segments, which shows that the dialog history is more important than those future expectations when deciding what to say. Still, the information that can be extracted from preceding and future segments is complementary, as revealed by the results when they are used in combination. In this case, we observed an improvement of 4.18 percentage points in relation to the scenario without context information on \ac{SwDA} and 0.69 on \ac{MRDA}. It is important to note that the 84.15\% achieved on \ac{SwDA} are in line with the inter-annotator agreement for the dataset.

Table~\ref{tab:resspeaker} shows the results achieved when using turn-taking information. We can see that this information is much less relevant than that provided by the classifications of the surrounding segments. Overall, the highest improvement is achieved when considering turn-taking information from the three preceding turns. In this case, we observed an improvement of 0.21 percentage points on \ac{SwDA} and 0.05 on \ac{MRDA} in relation to the scenario without context information. When only the first preceding segment was considered, as in the study by \shortciteA{Liu2017}, that is, whether the speaker of the current segment changed in relation to the previous segment, the improvement was not significant. Furthermore, since turn-taking information is represented as a sequence of flags, its summary is not able to provide additional information. However, since the summary of the turn-taking information of just three preceding segments achieved better results than that of the whole dialog history, we can assume that only the recent turn-taking information is relevant for the task.

\begin{table} [ht]
\begin{center}
    \begin{tabular}{l c c c c}
        \toprule
                                            &  \multicolumn{2}{c}{\textbf{SwDA}} &  \multicolumn{2}{c}{\textbf{MRDA}} \tabularnewline
        \textbf{Approach}                   & \textbf{$\mu$} & \textbf{$\sigma$} & \textbf{$\mu$} & \textbf{$\sigma$} \tabularnewline
        \midrule                    
        Without Context                     &          79.34 &              0.10 &          88.69 &              0.02 \tabularnewline
        \midrule              
        1 Preceding                         &          79.39 &              0.07 &          88.71 &              0.08 \tabularnewline
        3 Preceding Flat                    &          79.55 &              0.09 &          88.74 &              0.01 \tabularnewline
        Whole History Flat                  &          79.05 &              0.32 &          88.63 &              0.08 \tabularnewline
        3 Preceding Summary                 &          79.45 &              0.10 &          88.72 &              0.01 \tabularnewline
        Whole History Summary               &          79.35 &              0.11 &          88.70 &              0.05 \tabularnewline
        \bottomrule     
    \end{tabular}
\end{center}
\caption{Accuracy (\%) results using context information concerning turn taking.}
\label{tab:resspeaker}
\end{table}

Finally, Table~\ref{tab:rescontext} shows the results achieved using the best approaches for providing information from the surrounding segments and concerning the turn-taking history, as well as their combination. We can see that by combining turn-taking information with that provided by the classifications of the preceding segments, the performance improved in relation to the scenario without context information by 3.86 percentage points on \ac{SwDA} and 0.61 on \ac{MRDA}. This means that the turn-taking information improved the performance of the scenario with context information from the preceding segments by 0.44 percentage points on \ac{SwDA} and 0.11 on \ac{MRDA}. Thus, turn-taking information seems to be more informative when paired with the classifications of the corresponding segments. Finally, when considering the classifications of future segments as well, we achieved the highest performance in this study, with 84.40\% and 89.56\% accuracy on \ac{SwDA} and \ac{MRDA}, respectively. 

\begin{table} [ht]
\begin{center}
    \begin{tabular}{l c c c c}
        \toprule
                                                         &  \multicolumn{2}{c}{\textbf{SwDA}} &  \multicolumn{2}{c}{\textbf{MRDA}} \tabularnewline
        \textbf{Approach}                                & \textbf{$\mu$} & \textbf{$\sigma$} & \textbf{$\mu$} & \textbf{$\sigma$} \tabularnewline
        \midrule                                    
        Without Context                                  &          79.34 &              0.10 &          88.69 &              0.02 \tabularnewline
        \midrule                        
        Preceding Classifications                        &          82.76 &              0.13 &          89.19 &              0.01 \tabularnewline
        Preceding + Future Classifications               &          84.15 &              0.27 &          89.38 &              0.04 \tabularnewline
        \midrule
        Turn Taking                                      &          79.55 &              0.09 &          88.74 &              0.01 \tabularnewline
        \midrule
        Preceding Classifications + Turn Taking          &          83.20 &              0.14 &          89.30 &              0.04 \tabularnewline
        Preceding + Future Classifications + Turn Taking &          84.40 &              0.18 &          89.56 &              0.06 \tabularnewline
        \bottomrule     
    \end{tabular}
\end{center}
\caption{Accuracy (\%) results using context information from different sources.}
\label{tab:rescontext}
\end{table}

\subsection{Results on the Test Sets}

Table~\ref{tab:restest} shows the results achieved on the test sets of the two corpora. We report the results of the approach which considers context information from the preceding segments only, as well as of that which also considers future information. While the first simulates a scenario in which a dialog system interacts with its conversational partner, the latter simulates a posthumous annotation scenario. Furthermore, for the first scenario, we report the results achieved using the gold standard annotations of the preceding segments, which reveal the upper bound of the performance, as well as using the classifications predicted by the classifier itself, which provide an estimation of the performance in a real scenario.

\begin{table} [ht]
\begin{center}
    \begin{tabular}{l c c c c}
        \toprule
                                            &  \multicolumn{2}{c}{\textbf{SwDA}} &  \multicolumn{2}{c}{\textbf{MRDA}} \tabularnewline
        \textbf{Approach}                   & \textbf{$\mu$} & \textbf{$\sigma$} & \textbf{$\mu$} & \textbf{$\sigma$} \tabularnewline
        \midrule      
        Gold Standard Preceding Annotations &          80.49 &              0.10 &          90.93 &              0.05 \tabularnewline
        Automatic Preceding Classifications &          79.11 &              0.16 &          90.63 &              0.06 \tabularnewline
        Preceding + Future Classifications  &          82.34 &              0.20 &          91.27 &              0.05 \tabularnewline
        \bottomrule     
    \end{tabular}
\end{center}
\caption{Accuracy (\%) results on the test sets of both datasets.}
\label{tab:restest}
\end{table}

We can see that while on \ac{SwDA} the results on the test set are lower than on the validation set, that is not true on \ac{MRDA}. When considering gold standard annotations, on \ac{SwDA}, the accuracy is 2.71 percentage points lower when considering preceding segments only and 2.06 when considering future segments as well. On \ac{MRDA}, the improvements are of 1.63 and 1.71 percentage points, respectively.

The decrease in performance when using automatic classifications of the preceding segments is of 1.38 percentage points on \ac{SwDA} and 0.30 on \ac{MRDA}. The lower decrease on \ac{MRDA} is consistent with the lower influence of context information on the prediction of the dialog act classes of that dataset. On \ac{SwDA}, the performance decrease is lower than that observed in previous studies~\shortcite{Ribeiro2015,Liu2017}. This is due to the higher performance of the base classifier, which leads to results closer to the upper bound. 

\subsection{Comparison with Previous Approaches}

Of the studies described in Section~\ref{sec:related}, those by \shortciteA{Khanpour2016} and \shortciteA{Liu2017} are the ones reporting highest performance. Table~\ref{tab:rescomparison} compares the results of our approach with those of the approaches described in those studies. Since \shortciteA{Liu2017} evaluated their approach using a non-standard partition of \ac{SwDA}, we replicated their approach and applied it on the standard partition, as well as on \ac{MRDA}. Since we do not have access to the Word2Vec embeddings trained on Facebook data used in their experiments, we used the ones trained on Wikipedia provided by \shortciteA{Khanpour2016}. For completeness, and given the similar results we achieved using the \ac{CNN}- and \ac{RNN}-based segment representation approaches, we also replicated the approach by \shortciteA{Khanpour2016}.

\begin{table} [ht]
\begin{center}
    \begin{tabular}{l c c c c}
        \toprule
                                          &   \multicolumn{2}{c}{\textbf{SwDA}} &   \multicolumn{2}{c}{\textbf{MRDA}} \tabularnewline
        \textbf{Approach}                 & \textbf{Validation} & \textbf{Test} & \textbf{Validation} & \textbf{Test} \tabularnewline
        \midrule      
        \shortciteA{Khanpour2016}         &               76.86 &         73.75 &               77.65 &         78.30 \tabularnewline
        \shortciteA{Liu2017}              &               81.15 &         78.74 &               78.14 &         78.83 \tabularnewline
        \midrule         
        Our Approach (Preceding)          &               83.20 &         80.49 &               89.30 &         90.93 \tabularnewline
        Our Approach (Preceding + Future) &               84.40 &         82.34 &               89.56 &         91.27 \tabularnewline
        \bottomrule  
    \end{tabular}
\end{center}
\caption{Comparison with the performance of previous approaches. In this case, we only report the average accuracy (\%) over ten runs.}
\label{tab:rescomparison}
\end{table}

In their paper, \shortciteA{Khanpour2016} reported 73.9\% and 80.1\% accuracy on the validation and test sets of \ac{SwDA}, respectively, as well as 86.8\% accuracy on the test set of \ac{MRDA}. Although we replicated every step of the described approach, we were not able to achieve the same results, especially the high performance reported on the test sets. In fact, in all of our experiments on \ac{SwDA}, we were never able to achieve better results on the test set than on the validation set.

On the other hand, as expected, the approach by \shortciteA{Liu2017} performed better than the one by \shortciteA{Khanpour2016} in our experiments, since the latter does not consider context information.

Finally, we can see that our approach surpasses both of those approaches. Thus, we can state that it achieves the current state-of-the-art results on automatic dialog act recognition. Since we use the same segment representation approach as \shortciteA{Liu2017}, the major improvements come from the use of the combination of word- and character-level tokenization approaches, the use of BERT embeddings at the word level, and the summarized representation of context information. However, the latter is not as important when predicting the labels of \ac{MRDA} as when predicting those of \ac{SwDA}.

%
%
%
%

\section{Conclusions}
\label{sec:conclusions}

We explored multiple approaches on token, segment, and context information representation in the context of automatic dialog act recognition using \acp{DNN}. We performed experiments on \ac{SwDA} and \ac{MRDA}, which are two of the most explored corpora for the task. Overall, using the combination of the best approaches for each step, we were able to achieve results that surpassed the previous state-of-the-art on the task on both corpora.

In terms of token representation, we have explored approaches at the character, word, and functional levels, as well as their combination. Starting with the typically used word level, we have shown that, since the dialogs do not have fixed domain nor structure, using pre-trained embeddings leads to better performance than using a random initialization that adapts during the training phase. Furthermore, using dependency-based embeddings leads to better results than using representations generated using Word2Vec and FastText, since they include segment structure information that is relevant for identifying some dialog acts. However, the best results at the word level were achieved using contextualized embeddings, that is, individualizing the representation of each word by including information concerning the context in which it appears. Similarly to what happened on other \ac{NLP} tasks~\shortcite{Devlin2018}, using BERT embeddings led to better results than the simpler ELMo embeddings, especially on \ac{MRDA} where the difference was around 8 percentage points.

As we had previously shown~\shortcite{Ribeiro2018,Ribeiro2019}, our experiments at the character-level revealed that there is important information for the task at a sub-word level, which approaches that rely on word-level tokenization are not viable to capture. This information mostly concerns morphological aspects of the words, such as affixes and lemmas, which reveal their function and provide an important cue to identify intention. On its own, the character-level approach was able to achieve results similar to those achieved using pre-trained word embeddings on \ac{SwDA}. On \ac{MRDA}, the character-level approach surpassed even the results achieved using BERT embeddings. However, the character-level approach is also unable to capture all the information captured by the word-level approaches. Thus, the best results are achieved when both approaches are used in combination.

At the functional-level, our experiments have shown that using just the lemma instead of the whole word leads to better results on \ac{MRDA} and slightly worse on \ac{SwDA}. This suggests that affixes are only required for transmitting more specific aspects of intentions. By replacing the words with the corresponding \ac{POS} tags, the results diverged on the two corpora. While it was highly detrimental on \ac{SwDA}, decreasing performance by around 6.5 percentage points, it improved the performance on \ac{MRDA} by around 5 percentage points. Once again, these results are explained by the more abstract dialog acts used on \ac{MRDA}, which are more dependent on the function of the words and the structure of the segment than on the presence of specific words. Also, note that, since the embeddings of the \ac{POS} tags are initialized randomly, these results are in comparison with the scenario with randomly initialized word embeddings for fairness. If we consider the word-level approach using BERT embeddings on \ac{MRDA}, the results achieved using \ac{POS} tags are still surpassed. Still, on both corpora, the use of the fine-grained set of \ac{POS} tags, which includes morphological information, led to better results than using the coarse-grained set, which only reveals the word function. Finally, contrarily to what happened at the character-level, the combination of the functional-level with the word-level did not lead to improvement.

In terms of segment representation approaches at the word level, we compared the performance of the \ac{RNN}- and \ac{CNN}-based approaches used by \shortciteA{Khanpour2016} and \shortciteA{Liu2017}, respectively, as well as an \ac{RCNN}-based one that aimed at combining the advantages of the other two. Overall, the three approaches achieved similar results on both corpora. The \ac{RCNN}-based approach was not able to improve the results of the other two, since its advantage comes from the fact that it appends a summary of the context surrounding each token to its representation. Since we used BERT embeddings, which are already contextualized, that information is redundant. In fact, by reducing the segment representation approach to a simple max pooling operation applied on the BERT embeddings of the tokens, the decrease in performance was below a single percentage point on \ac{SwDA} and insignificant on \ac{MRDA}. Thus, we opted for using the \ac{CNN}-based approach, since it consumes fewer resources than the other two and it is also the one used at the character-level. However, in the latter case, the three context windows of three, five, and seven characters are selected to target specific morphological aspects. 

Concerning context information, we focused on that provided by preceding segments, since those are the ones available to a dialog system attempting to identify its conversational partner's intention. Previous dialog act recognition studies have shown that the best way to represent relevant context information from preceding segments is in the form of their classifications and not their words. However, in those studies, the sequentiality of the preceding segments, which is one of their main characteristics, was not appropriately represented. We approached this gap by reducing the representation of information from preceding segments to a summary of the sequence of their classifications, generated by a recurrent approach. This approach is able to capture not only information concerning sequentiality, but also relevant information from distant segments, as shown by the performance improvements in relation to when only considering the classification of three preceding segments and the flattened dialog history. However, the performance gains when using this context information are higher when predicting the labels of \ac{SwDA} than when predicting those of \ac{MRDA}, since the latter are less ambiguous. Additionally, on \ac{SwDA}, the difference between the upper bound of this approach, achieved when using the gold standard annotations, and the real performance, achieved when using the classifications predicted by the classifier itself was of 1.38 percentage points. This value is lower than those observed in previous studies~\shortcite{Ribeiro2015,Liu2017}, which proves the better performance of the base classifier.  

In their study, \shortciteA{Liu2017} have shown that speaker-change information in relation to the preceding segment is relevant for the task. As an extension, we explored the whole turn-taking history instead of just the relation with the preceding segment. To provide information concerning turn-taking we used approaches similar to those used for providing context information concerning the classification of preceding segments. However, in this case, since the turn-taking information is represented as a single flag, the summarization approach was not able to improve the results and the best results were achieved using a flat representation of turn-taking information in the three preceding segments.

Additionally, to simulate the annotation environment, in which the annotators have access to the whole dialog, we performed experiments that provided classification information from future segments in the same manner as that from preceding segments. In this scenario, when information concerning turn-taking was combined with that extracted from all surrounding segments, our approach surpassed the inter-annotator agreement of 84\% on the validation set of \ac{SwDA}, which means that our classifier has performance similar to that of a human annotator. However, on the test set, the performance was still 1.66 percentage points below that value.

Direct comparison with the results reported in previous studies was not straightforward, since \shortciteA{Liu2017} used a non-standard dataset partition to evaluate their experiments and \shortciteA{Khanpour2016} reported high discrepancies between the results on the validation and test sets of \ac{SwDA}. Thus, we replicated their experiments in order to directly compare the results with those of our approach. First, as expected, the approach by \shortciteA{Liu2017} performed better than that by \shortciteA{Khanpour2016}, since the latter does not use context information. Still, our approach was able to outperform it on the validation and test sets of both corpora. On \ac{SwDA}, the improvement is around 2 percentage points, while on \ac{MRDA} it is above 10 percentage points. Since we use the same segment representation, the major improvements come from the use of the combination of word- and character-level tokenization approaches, the use of BERT embeddings at the word level, and the summarized representation of context information.

In terms of future work, there are still some aspects that can be explored, especially in terms of token  and context information representation. Concerning token representation, we intend to assess whether there are specific word classes that can be abstracted by the corresponding \ac{POS} tags. In terms of segment representation, we believe that the room for improvement is reduced, since, at the word-level, BERT embeddings already capture most of the inter-token information that the segment representation approaches are able to capture. However, it would still be interesting to explore the influence of attentional mechanisms. Concerning context information, we intend to explore the use of other sources that may be relevant for the task, such as the domain and context of the dialog. Furthermore, in scenarios with more than two speakers, we intend to explore the use of speaker identifiers instead of flags to provide turn-taking information. Finally, it would be interesting to explore other approaches to combine the information extracted at different tokenization levels, as well as to merge context information with the segment representation.

%
%

%
%

\acks{This work was supported by national funds through \ac{FCT} with reference UID/CEC/50021/2019 and by Universidade de Lisboa.}

%
%

\vskip 0.2in
\bibliography{references}
\bibliographystyle{theapa}

\end{document}